\pdfoutput=1
\documentclass[final]{cvpr}
\usepackage{times}
\usepackage{epsfig}
\usepackage{graphicx}
\usepackage{amsmath}
\usepackage{amssymb}
\usepackage{algorithm}
\usepackage{listings}
\usepackage{multirow}
\usepackage[table,xcdraw]{xcolor}
\usepackage{booktabs}
\usepackage{verbatim}
\usepackage{color, colortbl}
\usepackage{bm}
\usepackage{enumitem}
\usepackage{pifont}
\usepackage{subfigure}
\usepackage[accsupp]{axessibility}

\newcommand{\cmark}{\ding{51}}%
\newcommand{\xmark}{\ding{55}}%
\newcommand\minisection[1]{\vspace{1mm}\noindent \textbf{#1}}
\definecolor{Gray}{gray}{0.9}

\usepackage[font=small,labelfont=small]{caption}

\usepackage{array}

\newcolumntype{x}[1]{>{\centering\let\newline\\\arraybackslash\hspace{0pt}}p{#1}}

\usepackage[pagebackref=true,breaklinks=true,colorlinks,bookmarks=false]{hyperref}

\def\cvprPaperID{6196} % *** Enter the ICCV Paper ID here
\def\confYear{ICCV 2021}

\begin{document}

%%%%%%%%% TITLE
\title{DeFRCN: Decoupled Faster R-CNN for Few-Shot Object Detection}

\author{Limeng Qiao   \quad 
	Yuxuan Zhao  \quad 
	Zhiyuan Li         \quad 
	Xi Qiu\thanks{\textit{Corresponding author}.} \quad 
	Jianan Wu        \quad 
	Chi Zhang         \\
	Megvii Technology   \\
	{\tt\small \{qiaolimeng, zhaoyuxuan, lizhiyuan, qiuxi, wjn, zhangchi\}@megvii.com}
}

\maketitle

%%%%%%%%% ABSTRACT
\begin{abstract}   % \vspace{-0.5em}
	Few-shot object detection, which aims at detecting novel objects rapidly from extremely few annotated examples of previously unseen classes, has attracted significant research interest in the community. Most existing approaches employ the Faster R-CNN as basic detection framework, yet, due to the lack of tailored considerations for data-scarce scenario, their performance is often not satisfactory. In this paper, we look closely into the conventional Faster R-CNN and analyze its contradictions from two orthogonal perspectives, namely multi-stage (RPN vs. RCNN) and multi-task (classification vs. localization). To resolve these issues, we propose a simple yet effective architecture, named Decoupled Faster R-CNN (DeFRCN). To be concrete, we extend Faster R-CNN by introducing Gradient Decoupled Layer for multi-stage decoupling and Prototypical Calibration Block for multi-task decoupling.  The former is a novel deep layer with redefining the feature-forward operation and gradient-backward operation for decoupling its subsequent layer and preceding layer, and the latter is an offline prototype-based classification model with taking the proposals from detector as input and boosting the original classification scores with additional pairwise scores for calibration. Extensive experiments on multiple benchmarks show our framework is remarkably superior to other existing approaches and establishes a new state-of-the-art in few-shot literature \footnote{\footnotesize\url{https://github.com/er-muyue/DeFRCN}}.
\end{abstract}

%%%%%%%%% BODY TEXT
\vspace{-0.4cm}
\section{Introduction}
Recently, deep neural networks have achieved state-of-the-art on a variety of visual tasks, $\eg$ image classification \cite{ dai2017deformable, he2016deep, huang2017densely} and object detection \cite{cai2018cascade, dai2016r, girshick2015fast, girshick2014rich, lin2017feature, redmon2016you, redmon2017yolo9000, ren2015faster}. However, these leaps of performance arrive only when a large amount of annotated data is available.  Since it is often labor-intensive to obtain adequate labelled data, the number of available samples severely limits the applications of current vision systems. Besides, compared to the ability of human to quickly extract novel concepts from few examples, these deep models are still far from satisfactory. 

\begin{figure}[t]
	\vspace{0.1cm}
	\begin{center}
		\includegraphics[width=0.9\linewidth]{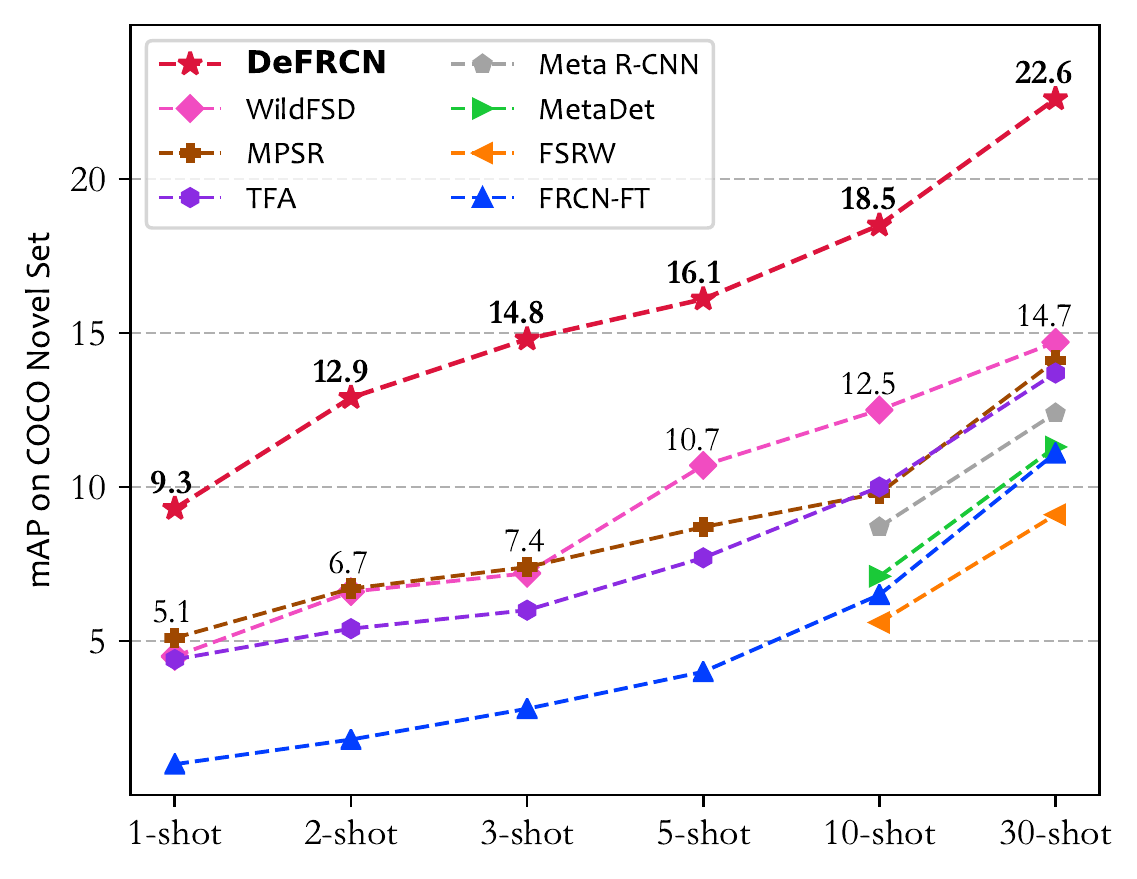}
	\end{center}
	\vspace{-0.5cm}
	\caption{FSOD performance (mAP) on COCO \cite{lin2014microsoft} novel set at different shot numbers. The proposed DeFRCN is remarkably superior to other state-of-the-art approaches.}
	\vspace{-0.6cm}
	\label{fig:scores}
\end{figure}

It is thus of attracting major research interest on few-shot learning \cite{chen2019closer, koch2015siamese, li2017meta, qiao2019transductive, Snell2017, Sung2018, vinyals2016matching}, which employs the idea of learning novel concepts rapidly and generalizing well in data-scarce scenario. As one of the research branches, few-shot object detection (FSOD) is a much more challenging task than both few-shot classification and object detection \cite{chen2018lstd, kang2019few, wang2020frustratingly, Xiao2020FSDetView,yan2019meta}. At present, most FSOD approaches prefer to follow the meta-learning paradigm to acquire more task-level knowledge and generalize better to novel classes. However, these methods usually suffer from a complicated training process and data organization,  which results in limited application scenarios. In contrast, the finetune-based methods that exist as another research branch of FSOD,  are very simple and efficient \cite{wang2020frustratingly}. By adopting a two-stage fine-tuning scheme, this series is comparable to meta methods. Yet, due to most parameters are pre-trained on base domain and then frozen on novel set, they may fall down the severe shift in data distribution and underutilization of novel data. 

\begin{figure}[t]
	\vspace{-0.cm}
	\begin{center}
		\includegraphics[width=1.0\linewidth]{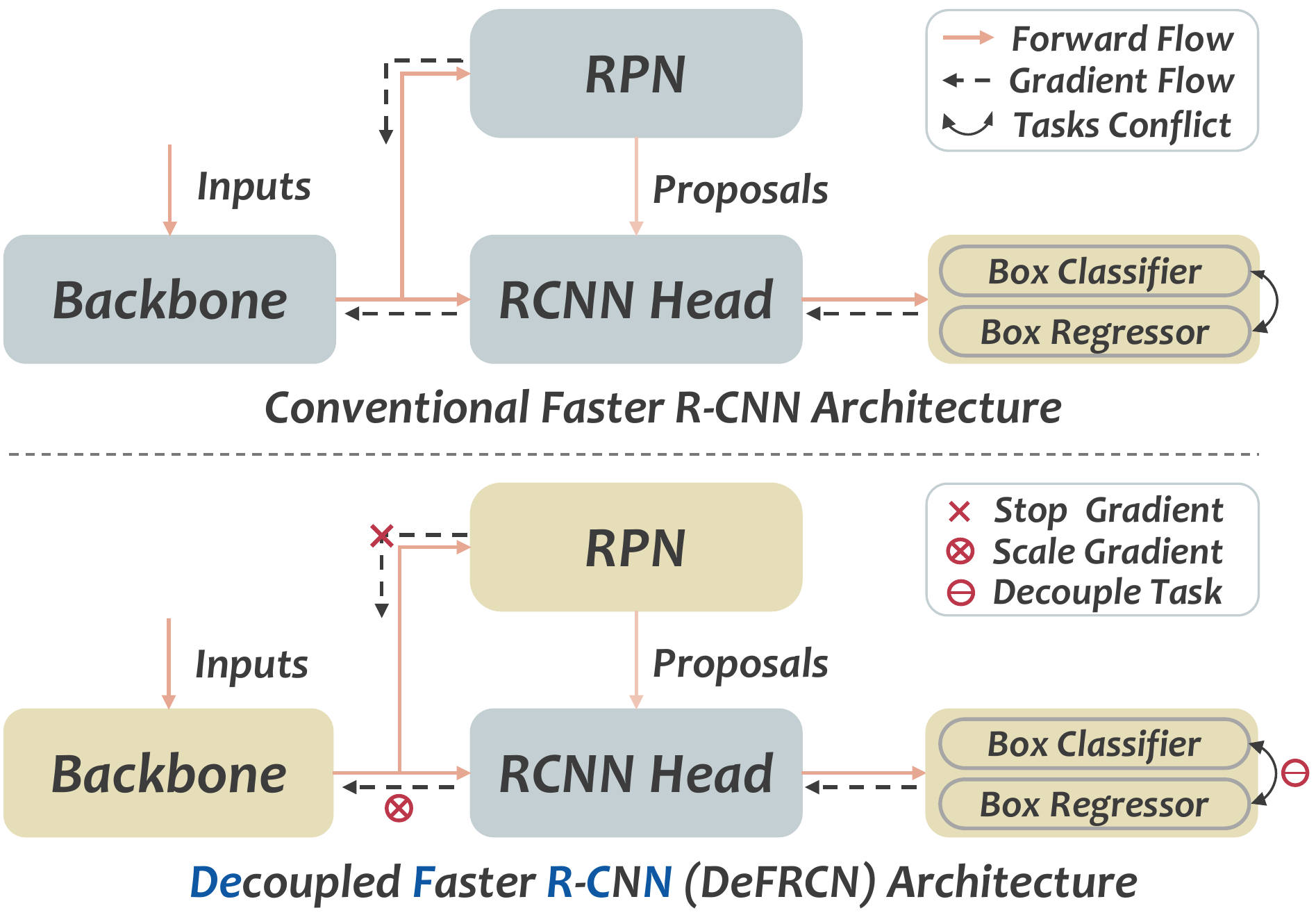}
	\end{center}
	\vspace{-0.55cm}
	\caption{Comparison of Faster R-CNN and our motivation. We performs \textbf{\textit{stop-gradient}} between RPN and backbone, meanwhile, \textbf{\textit{scale-gradient}} between RCNN and backbone, as well as decouple conflict tasks between classifier and regressor. The yellow blocks are trainable during fine-tuning.}
	\vspace{-0.58cm}
	\label{fig:motivation}
\end{figure}

Regardless of the meta-based or finetune-based method, Faster R-CNN \cite{ren2015faster} has been widely used as the basic detector and achieved good performance. However, its original architecture is designed for conventional detection and lacks of tailored consideration for few-shot scenario, which limits the upper bound of existing approaches. Concretely, on the one hand, as a classic two-stage stacking architecture, ($\ie$, backbone, RPN and RCNN, see Fig.\ref{fig:motivation}), Faster R-CNN may encounter  an intractable conflict when it performs joint optimization end-to-end between \textbf{\textit{class-agnostic}} RPN and \textbf{\textit{class-relevant}} RCNN through the shared backbone. On the other hand, as a multi-task learning paradigm ($\ie$, classification and localization), RCNN needs translation-invariant features for box classifier whereas translation-covariant features for box regressor. These mismatched goals potentially generate so many low-quality scores and then further lead to the reduced classification power. Moreover, since there are only a few samples available during learning, these above contradictions will be further exacerbated. 

Motivated by the above observations, we extend Faster R-CNN for few-shot scenario from two orthogonal perspectives: (1) multi-stage view. As shown in Fig.\ref{fig:motivation}, the Faster R-CNN contains three components, $\ie$, backbone, RPN and RCNN, which interact with each other through feature-forward and gradient-backward. Due to the contradiction mentioned above between RPN and RCNN, we present to alleviate the entire model from being dominated by one of them with tailoring the degree of decoupling between three modules through gradient. (2) multi-task view. The task conflict between classification and regression affects the quality of features, which in turn damages the performance of box head outputs, $\ie$, category scores and box coordinates. We employ an efficient score calibration module only on the classification branch to achieve the purpose of decoupling the above two tasks.

\begin{figure*}
	\vspace{-0.cm}
	\begin{center}
		\includegraphics[width=1.0\linewidth]{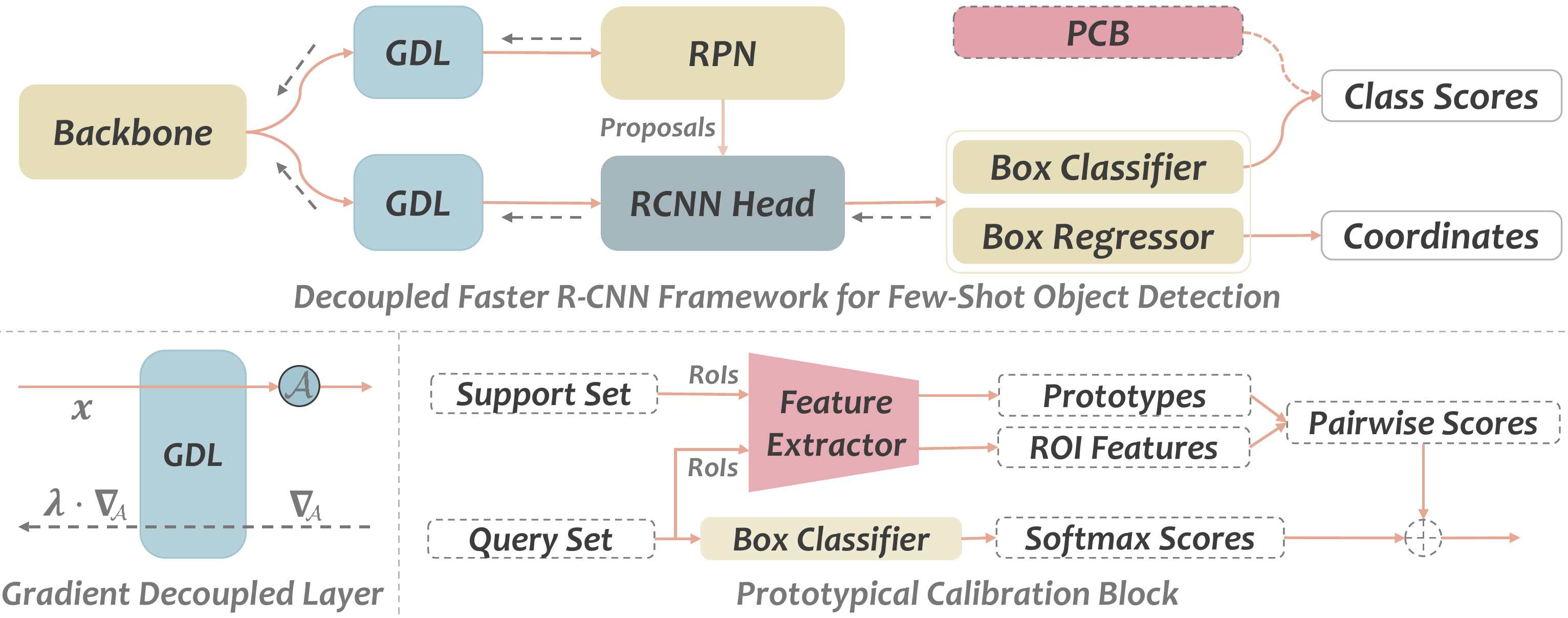}
	\end{center}
	\vspace{-0.45cm}
	\caption{\textbf{The architecture of Decoupled Faster R-CNN (DeFRCN) for few-shot object detection}. Compared to the standard Faster R-CNN, there are two Gradient Decoupled Layers (sky-blue) and an offline Prototypical Calibration Block (red) are inserted into the framework to perform decoupling for multi-stage and multi-task,  respectively. The $\mathcal{A}$ is the affine transformation layer in GDL and $\oplus$ is score fusion operation in PCB. Moreover, yellow and dark-blue indicate that the block is trainable and frozen during fine-tuning. The orange solid and black dotted lines represent forward flow and gradient flow.}
	\vspace{-0.26cm}
	\label{fig:framework}
\end{figure*}

This paper proposes a simple yet effective approach, named Decoupled Faster R-CNN (DeFRCN), to perform both multi-stage decoupling and multi-task decoupling for few-shot object detection. The overall architecture is very straightforward as demonstrated in Fig.\ref{fig:framework}. Compared to the standard Faster R-CNN \cite{ren2015faster}, DeFRCN additionally contains two Gradient Decoupled Layer (GDL) and an offline Prototypical Calibration Block (PCB).  The former ones are inserted between the shared backbone and RPN, meanwhile, between the backbone and RCNN to adjust the degree of decoupling among three modules, and the latter is parallel to the box classifier for further score calibration. Specifically, during the forward-backward propagation, GDL performs a learnable affine transformation on the forward feature maps and simply multiplies the backward gradient by a constant, which decouples the subsequent module and preceding module efficiently. Moreover, PCB is initially equipped with a well pre-trained classification model ($\eg$ ImageNet Pretrain) and a set of novel support prototypes. Then it takes the region proposals from few-shot detector as input and boosts the original softmax scores with additional prototype-based pairwise scores. As an interesting by-product, we find that just adopting PCB only in the inference phase can greatly improve the performance of few-shot detectors, with no extra training effort, which makes the PCB data-efficient and plug-and-play.

\noindent The main contributions of our approach are three-folds:
\begin{itemize}[leftmargin=12pt,topsep=2pt, parsep=-1pt]
	
	\item We look closely into the conventional Faster R-CNN and propose a simple yet effective architecture for few-shot detection, named Decoupled Faster R-CNN, which can be learned end-to-end via straightforward fine-tuning.
	
	\item To deal with the data-scarce scenario, we further present two novel modules, \ie GDL and PCB, to perform decoupling among multiple components of Faster R-CNN and boost classification performance respectively.
	
	\item DeFRCN is remarkably superior to SOTAs on various benchmarks, revealing the effectiveness of our approach.
	
\end{itemize}

%----------------------------------------------------------------------------------

\section{Related Work}

\subsection{General Object Detection}
General object detection based on deep neural networks are currently divided into two main branches, $\ie$, two-stage proposal-based paradigm \cite{cai2018cascade, dai2016r, girshick2015fast, girshick2014rich, he2017mask,  lin2017feature,ren2015faster} and one-stage proposal-free one \cite{bochkovskiy2020yolov4, liu2016ssd, redmon2016you, redmon2017yolo9000, redmon2018yolov3}, which both have witnessed fantastic progress on numerous large-scale benchmarks. The R-CNN series falls into the former line of work, which firstly generates a set of potential objects with region proposal network (RPN) \cite{ren2015faster} and then performs category classification and box localization for end-to-end detection. In contrast, one-stage detectors endeavour to directly produce final predictions from the feature map without RPN module, usually have the advantages of inference speed but the detection performance is often not as good as two-stage approaches. However, all these frameworks uniformly assume that a large amount of annotated data from seen domain can be accessed, which may be stuck in troubles in data-scarce scenarios or novel unseen domains.

\subsection{Few-Shot Learning}
Few-shot learning, which aims at learning to learn general knowledge slowly from abundant base data and extracting novel concepts rapidly from extremely few examples of new-coming classes, has been recently featured into the meta-learning based \cite{vilalta2002perspective} and fine-tuning based \cite{pan2009survey} paradigms. As a recognition case of few-shot learning, few-shot classification has been widely investigated until now. In the literature, a large amount of studies that follow the idea of meta-learning to alleviate severe over-fitting can be divided into two streams, namely, optimization approaches \cite{andrychowicz2016learning, finn2017model, li2017meta, nichol2018first, qiao2018few, ravi2016optimization} and metric approaches \cite{koch2015siamese, qiao2019transductive, Snell2017, Sung2018, vinyals2016matching}. The former intents to learn efficient parameter updating rules \cite{ravi2016optimization} or good parameters initialization strategies  \cite{finn2017model}, and the latter focuses on obtaining a generalizable embedding metric space to perform pairwise similarity of inputs. In addition to meta-based approaches, some simple fine-tuning based methods \cite{chen2019closer, tian2020rethinking} are attaching more and more attention in the few-shot community. These methods show that just fine-tuning a linear classifier on top of a pre-trained model surprisingly achieves competitive performance with the meta-based approaches. Compared to classification, the solutions for other tasks, such as object detection and segmentation, are still underdeveloped.

\subsection{Few-Shot Object Detection}
Since previous detectors usually require a large amount of annotated data, few-shot detection has attracted more and more interest recently \cite{bansal2018zero, dong2018few,fan2020fgn, perez2020incremental, rahman2020any, wang2019few, wang2019meta, xiao2021few, yang2020context}.
Similar to classification task \cite{Snell2017, sun2019meta}, most of the current few-shot detectors focus on the meta-learning paradigm. FSRW \cite{kang2019few} is a light-weight meta-model based on YOLOv2 \cite{redmon2017yolo9000} to re-weight the importance of features with channel-wise attention, and then adapt these features to promote novel detection. Yet, instead of employing attention on the whole feature map, Meta R-CNN \cite{yan2019meta} focuses on the attention of each RoI feature. Furthermore, FSDView \cite{Xiao2020FSDetView} puts forward a novel feature aggregation scheme, which leverages on base classes feature information to improve the performance on novel classes. From the perspective of attention on RPN, FSOD \cite{fan2020few} utilizes support information to filter out most background boxes and those in non-matching categories. Although meta-based approaches have been extensively studied recently, there are still some other meta-free methods. RepMet \cite{karlinsky2019repmet} incorporates a modified prototypical network as classification head into a standard object detector. And TFA \cite{wang2020frustratingly}  proposes a simple approach based on transfer learning, that only fine-tunes the last layer of existing detectors on rare classes, which are comparable to the previous meta-based methods. Instead, our approach, which  also follows the idea of fine-tuning, jointly trains the almost entire detector with novel gradient decoupled layer and prototypical calibration block, outperforming all above meta-based and finetune-based  approaches.

%-------------------------------------------------------------------------
\section{Methods}

In this section, we first introduce the setup of few-shot object detection in Section \ref{sec:problem-setting}. Then we revisit conventional Faster R-CNN in Section \ref{sec:faster-rcnn} and elaborate our Decoupled Faster R-CNN (DeFRCN) in Section \ref{sec:decouple-rcnn}. 

\subsection{Problem Setting} \label{sec:problem-setting}
As in various previous work \cite{fan2020few, kang2019few, wang2020frustratingly, Xiao2020FSDetView}, we follow the standard problem settings of few-shot object detection in our paper. Specifically, the whole learning procedure is organized into the form of two-stage fine-tuning  paradigm, which gradually collects transferable knowledge across a large base set $\mathcal{D}_{\text{\textit{base}}}$ with abundant annotated instances and  performs adaptation quickly on novel support set $\mathcal{D}_{\text{\textit{novel}}}$ with only a few samples per category. Note that the base classes $\mathcal{C}_{\text{\textit{base}}}$ in  $\mathcal{D}_{\text{\textit{base}}}$ and the novel classes $\mathcal{C}_{\text{\textit{novel}}}$ in $\mathcal{D}_{\text{\textit{novel}}}$ are non-overlapping, namely, $\mathcal{C}_{\text {\textit{base}}} \cap \mathcal{C}_{\text {\textit{novel}}}=\varnothing$. Given a sample $(x, y) \in \mathcal{D}_{\text {\textit{base}}} \cup \mathcal{D}_{\text {\textit{novel}}} $, where $x=\left\{o_i, i=1, ..., N\right\}$ is the input image with $N$ objects and $y=\left\{(c_i, b_i), i=1, ..., N \right\}$ denotes the category $c_i \in \mathcal{C}_{\text {\textit{base}}} \cup \mathcal{C}_{\text {\textit{novel}}} $ and the structured box annotations  $b_i$. Under this setting, the ultimate goal of our algorithm is to optimize a robust detector $\mathcal{F}$  based on the  $\mathcal{D}_{\text{\textit{base}}}$ and $\mathcal{D}_{\text{\textit{novel}}}$, then classify and localize unlabelled objects of a novel query set $\mathcal{D}_{\text{\textit{query}}}$  with classes $ \mathcal{C}_{\text {\textit{query}}}$, where $ \mathcal{C}_{\text {\textit{query}}} \subseteq \mathcal{C}_{\text {\textit{base}}} \cup \mathcal{C}_{\text {\textit{novel}}} $. The overall procedure, which follows the standard transfer learning,  can be summarized as follow,
\begin{equation} \label{eq:finetune-paradigm}
\mathcal{F}_{init}    \stackrel{\mathcal{D}_{base}}{\rightsquigarrow\rightsquigarrow} 
\mathcal{F}_{base} \stackrel{\mathcal{D}_{novel}}{\rightsquigarrow\rightsquigarrow}
\mathcal{F}_{novel}
\end{equation}
where $\mathcal{F}_{\text {\textit{init}}}$, $\mathcal{F}_{\text {\textit{base}}}$ and $\mathcal{F}_{\text {\textit{novel}}}$ denote the learned detectors in initialization, base training and novel fine-tuning stages respectively. The symbol $\rightsquigarrow$ indicates model training.

\subsection{Revisiting Faster R-CNN}  \label{sec:faster-rcnn}
As a two-stage stacking architecture, Faster R-CNN \cite{ren2015faster} consists of three function-detached modules for end-to-end training, $\ie$, a shared convolutional backbone for extracting \textbf{\textit{generalized}} features, an efficient Region Proposal Network (RPN) for generating \textbf{\textit{class-agnostic}} proposals and a task-specific RCNN head \cite{girshick2015fast} for performing  \textbf{\textit{class-relevant}} classification and localization. The whole learning procedure is illustrated in Fig.\ref{fig:motivation} (a). Concretely, the input image is first fed into the backbone to generate a high-level feature map, and then parallelly provided to the next two modules, $\ie$, RPN and RCNN. Second, with classifying and regressing a group of scale varying anchors of the feature map simultaneously, RPN generates a sparse set of high-quality region proposals. Finally,  on top of the shared feature map and proposals, RCNN pools each region-of-interest into a fixed size feature map with RoI pooling \cite{he2017mask},  and then performs box classifier and regressor for computing  the object category probabilities and fine-tuning the box boundaries respectively. All these modules are jointly optimized end-to-end by minimizing an unify objective function, which follows the multi-task learning paradigm as: 
\begin{equation} \label{eq:Faster-RCNN-Loss}
\mathcal{L}_{total} = (\underbrace{\mathcal{L}_{rpn}^{cls} +  \mathcal{L}_{rpn}^{reg}}_{rpn \  task}) \ + \ \eta \cdot (\underbrace{ \mathcal{L}_{rcnn}^{cls} + \mathcal{L}_{rcnn}^{reg}}_{rcnn \  task})
\end{equation}
where $\eta$ is a balanced hyper-parameter for different tasks.

\minisection{Problem of multi-task learning.}
It can be seen that  the above-mentioned three modules of Faster R-CNN constitute an unified multi-task learning (MTL) framework, yet there is a certain inconsistency among the optimization goals of these sub-networks. Specifically, with utilizing the feature maps extracted from hard-parameter shared \cite{vandenhende2021multi} backbone, RPN aims at generating class-agnostic region proposals to tell the network \textit{where to look}, while RCNN targets to perform region-based detection category by category to identify \textit{what to look}. Furthermore, the classification head needs translation invariant features whereas the localization head needs translation covariant features on the contrary. In spite of multi-task learning generally helps to improve the end-to-end performance of object detection as shown in Faster R-CNN \cite{ren2015faster}, the joint optimization with the Eq.\ref{eq:Faster-RCNN-Loss} may lead to possible suboptimal solution on individual tasks in order to balance the mismatched goals of them \cite{cheng2018revisiting, wu2020rethinking}.

\minisection{Problem of shared backbone.}
According to the arguments in \cite{ren2015faster}, the ultimate goal of shared backbone is to extract general features that are as suitable as possible for all downstream tasks. In fact, from the perspective of gradient flow in Fig.\ref{fig:motivation} (a), RPN and RCNN mutually exchange information of optimization through the shared backbone. However, due to the potential contradictions between RPN and RCNN, we notice that the current architecture may lead to the reduced few-shot detection power of the entire framework. Moreover, following the setting of Eq.\ref{eq:finetune-paradigm}, the shared backbone of few-shot novel detector $\mathcal{F}_{novel}$ is usually fine-tuned from a base domain detector $\mathcal{F}_{base}$.  During this two-stage cross-domain procedure, RPN may suffer from the foreground-background confusion, which means a proposal that belongs to background in the base training phase is likely to be foreground in the novel fine-tuning phase. Through the gradient from RPN, the shared convolutional layers propagate the tendency of over-fitting on base classes to backbone and RCNN. Although this is one of the convergence schemes to behave well on base domain, it potentially damages the ability to transfer to the novel set quickly and efficiently, especially in the data-scarce scenario. 

\subsection{Decoupled Faster R-CNN}  \label{sec:decouple-rcnn}

Motivated by the above arguments, we propose a simple yet effective approach, named Decoupled Faster R-CNN (DeFRCN), to tap into more potential of Faster R-CNN styled detectors in few-shot literature. Based on the idea of decoupling three functional modules ( $\ie$, backbone, RPN and RCNN) and two kinds of tasks ($\ie$, classification and localization), the overall architecture of our  method is very straightforward as demonstrated in Fig.\ref{fig:framework}, which has two Gradient Decoupled Layers (GDL) to adjust the degree of decoupling among three modules and an offline Prototypical Calibration Block (PCB) to improve the classification power of RCNN during the inference phase.

\subsubsection{Gradient Decoupled Layer}

In this section, we look into a different aspect of network design - how to customize the relationship between the upstream and downstream modules of the model. From the perspective of feature-forward and gradient-backward,  we introduce a novel architectural unit, denoted as the Gradient Decoupled Layer (GDL). During the forward propagation, GDL employs an affine transformation layer $\mathcal{A}$, which is parameterized by learnable channel-wise weights $\omega$ and bias $b$, to simply enhance feature representations and perform forward-decoupling. During the backward propagation, GDL takes the gradient from the subsequent layer, multiplies it by a constant $\lambda \in [0, 1]$ and passes it to the preceding layer, as illustrated in Fig.\ref{fig:framework}. Concretely, along with the back-propagation process passes through the GDL, the partial derivatives of the loss $\mathcal{L}_d$  that is downstream of the GDL with respect to the layer parameters $\theta_u$ that are upstream of the GDL get multiplied by $\lambda$, $\ie$, $\frac{\partial \mathcal{L}_d}{\partial \theta_u}$ (denoted as $\nabla$ in Eq.\ref{eq:gdl-backward}) is simply replaced with $\lambda \frac{\partial \mathcal{L}_d}{\partial \theta_u}$. Mathematically, we can formally treat GDL as a pseudo-function $\mathbb{G}_{(\mathcal{A}, \lambda)}$ defined by two equations describing its forward- and backward-propagation behaviour as follows:
\begin{align}
\mathbb{G}_{(\mathcal{A}, \lambda)}(x)                                                        &= \mathcal{A}(x) \label{eq:gdl-forward} \\
\frac{\mathrm{d} \mathbb{G}_{(\mathcal{A}, \lambda)}}{\mathrm{d} x} &= \lambda \nabla_{\mathcal{A}}  \label{eq:gdl-backward}
\end{align}
where $\mathcal{A}$ is an affine transformation layer, $\lambda \in [0, 1]$ is a decoupling coefficient and $\nabla_{\mathcal{A}}$ is the Jacobian matrix from the affine layer. In general, implementing such layer with existing deep learning frameworks are extremely simple, as defining procedures for forwardprop (affine transformation) and backprop (multiplying by a constant) is trivial. We provide the pseudo-code of GDL in Algorithm \ref{alg:gdl-code}.

\minisection{Perform Decoupling with GDL.}
Given a standard Faster R-CNN \cite{ren2015faster}, two GDLs are respectively inserted between the shared backbone and RPN ($\ie$, $\mathbb{G}_{rpn}$), as well as the shared backbone and RCNN ($\ie$, $\mathbb{G}_{rcnn}$),  which brings the part of DeFRCN architecture depicted in Fig.\ref{fig:framework}.  Specifically, during the forward propagation, the feature from shared backbone is transformed into different feature spaces through $\mathcal{A}_{rpn}$ and $\mathcal{A}_{rcnn}$. Moreover, during the backward propagation, we adjust the decoupling degree of three modules ($\ie$, backbone, RPN and RCNN) by applying different $\lambda_{rpn}$ and $\lambda_{rcnn}$ on gradients. More formally, we consider the following loss function with two separate GDLs as:
\begin{align}
\mathcal{L} = \ & \mathcal{L}_{rpn}\left( F_{rpn}(\bm {\mathbb{G}_{rpn}}(F_{b}(x; \theta_{b})); \theta_{rpn}), y_{rpn}\right) + \eta \ \cdot \ \nonumber\\
& \mathcal{L}_{rcnn}\left( F_{rcnn}(\bm {\mathbb{G}_{rcnn}}(F_{b}(x; \theta_{b})); \theta_{rcnn}), y_{rcnn}\right)
\label{eq:loss-with-gdl}
\end{align}
Here, $\bm {\mathbb{G}_{\cdot}}$  is the Gradient Decoupled Layer we proposed in this section, $\theta_b$, $\theta_{rpn}$ and $\theta_{rcnn}$ are learnable parameters for the backbone, RPN and RCNN respectively. Moreover, $\eta$ is a hyper-parameter to control the trade-off between $ \mathcal{L}_{rpn}$ and $\mathcal{L}_{rcnn}$ (usually is set to $1$).

\minisection{Optimization with GDL.}
Consistent with the optimization goal of Faster R-CNN, we seek the optimal parameters $\theta_b$, $\theta_{rpn}$ and $\theta_{rcnn}$, denoted as $\Theta$, for the function Eq.\ref{eq:loss-with-gdl} as:
\begin{align}
\Theta = \arg \min_{\Theta} \frac{1}{\mathcal{N}}\sum_{i=1}^\mathcal{N} \mathcal{L}, \ \ \Theta = \{ \theta_{b}, \theta_{rpn}, \theta_{rcnn} \}
\label{eq:gdl-solution}
\end{align}
where $\mathcal{N}$ is the number of training samples, and $\mathcal{L}$ is from the Eq.\ref{eq:loss-with-gdl}. Concretely, a gradient descent step can be described as:
\begin{align}
\theta_b          \quad & \leftarrow \quad \theta_b \;-\; \gamma \left(\lambda_{1} \frac{\partial \mathcal{L}_{rpn}}{\partial \theta_b} + \lambda_{2} \frac{\partial \mathcal{L}_{rcnn}}{\partial \theta_b} \right) \label{eq:upd-backbone} \\
\theta_{rpn}   \quad & \leftarrow \qquad \theta_{rpn} \; -\; \gamma \frac{\partial \mathcal{L}_{rpn}}{\partial \theta_{rpn}}\label{eq:upd-rpn}\\
\theta_{rcnn} \quad & \leftarrow \qquad \theta_{rcnn} \;-\; \gamma \frac{\partial \mathcal{L}_{rcnn}}{\partial \theta_{rcnn}}\label{eq:upd-rcnn}
\end{align}
where $\gamma$ is the learning rate, $\lambda_1$ and $\lambda_2$ are decoupling coefficients for RPN and RCNN respectively. It can be seen from Eq.\ref{eq:upd-rpn} and Eq.\ref{eq:upd-rcnn} that adding GDL does not affect the optimization of RPN and RCNN. However, the parameter update of sharing backbone is deeply affected by GDL in Eq.\ref{eq:upd-backbone}. We mainly analyze three important situations: (1) $\lambda_{1} = 0$ (or $\lambda_{2}=0$),  it is equivalent to \textbf{\textit{stopping gradient}} from RPN (or RCNN), and the update of $\theta_{b}$ will only be dominated by RCNN (or RPN); (2) $\lambda_{1} \in \left(0, 1\right]$ (or $\lambda_{2}  \in \left(0, 1\right]$), it is equivalent to \textbf{\textit{scaling gradient}}	from RPN (or RCNN), which means that the RPN (or RCNN) has individual contributions to the update of shared backbone; (3) $\lambda_{1} = \lambda_{2}=\tilde{\lambda}$, which is equivalent to multiplying the learning rate $\gamma$ of backbone by a small coefficient, $\ie$, $\tilde{\lambda}$, ensures that the update speed of $\theta_{b}$ is slower than $\theta_{rpn}$ and $\theta_{rcnn}$. Note that $\lambda < 0$ is meaningless for detection and more discussion  about $\lambda$ is mentioned in the supplementary material.

\begin{algorithm}
\caption{Gradient Decoupled Layer, PyTorch-like}
\label{alg:gdl-code}
\definecolor{codeblue}{rgb}{0.25,0.5,0.5}
\definecolor{codekw}{rgb}{0.85, 0.18, 0.50}
\lstset{
	backgroundcolor=\color{white},
	basicstyle=\fontsize{8.2pt}{8.2pt}\ttfamily\selectfont,
	columns=fullflexible,
	breaklines=true,
	captionpos=b,
	commentstyle=\fontsize{8.2pt}{8.2pt}\color{codeblue},
	keywordstyle=\fontsize{8.2pt}{8.2pt}\color{codekw},
}
\begin{lstlisting}[language=python]
# A: learnable channel-wise affine layer
# _lambda: gradient decoupling coefficient
	
class GradientDecoupledLayer(Function):
	
   # feature forward
   def forward(ctx, x, A, _lambda): 
      ctx._lambda = _lambda
      x = A(x)
      return x.view_as(x)
	
   # gradient backward
   def backward(ctx, grad_output):
      grad_output = grad_output * ctx._lambda
      return grad_output, None, None
	
   def decouple_layer(x, A, _lambda):
      return GradientDecoupleLayer(x, A, _lambda)   
	\end{lstlisting}
\end{algorithm}
%##################################################################################################

\subsubsection{Prototypical Calibration Block}
In this section, we introduce a novel metric-based score refinement module, termed as Prototypical Calibration Block (PCB),  to effectively decouple the classification and localization tasks during the inference time. In general, most of detectors parallelly deploy a classifier and a regressor on top of the shared network. However, classification needs translation invariant features whereas localization needs translation covariant features. Thus the localization branch may force the backbone to gradually learn translation covariant property, which potentially downgrades the performance of classifier. Due to model complexity, the extreme lack of annotated samples will further exacerbate this contradiction.

We notice that the under-explored few-shot classification branch generates a large amount of low-quality scores, which motivates us to eliminate high-scored false positives and remedy low-scored missing samples by introducing a Prototypical Calibration Block (PCB) for score refinement.
The overall pipeline is illustrated in Fig.\ref{fig:framework} (c). Concretely, our PCB consists of a strong classifier from ImageNet pre-trained model, a RoIAlign layer and a prototype bank.  Given a $M$-way $K$-shot task with support set $\mathcal{S}$, the PCB first extracts original image feature map and then employs RoIAlign with ground-truth boxes to produce $MK$ instance representations. Based on these features, we shrink the support set $\mathcal{S}$ to a prototype bank $\mathcal{P}=\{p_c\}_{c=1}^{M}$ with Eq.\ref{eq:prototype}:
\begin{align}
p_c = \frac{1}{\vert \mathcal{S}_c \vert} \mathop{\sum}\nolimits_{(x_i, \ y_i) \in \mathcal{S}_c} x_i \label{eq:prototype}
\end{align}
where $\mathcal{S}_c$ is a subset which contains samples with the same label $c$ in $\mathcal{S}$.
Given an object proposal $\hat{y_i} =  (c_i, s_i, b_i)$ produced by fine-tuned few-shot detector, where $b_i$ is the box boundaries, $c_i$ is the predicted category and $s_i$ is the corresponding score, PCB first performs RoIAlign on predicted box $b_i$ to generate object feature $x_i$, and then calculate the cosine similarity $s_i^{cos}$ between $x_i$ and $p_{c_i}$ as:
\begin{align}
s_i^{cos} = \frac{ x_i \cdot p_{c_i}}{\|x_i\| \|p_{c_i}\|}  \label{eq:cos_sim}
\end{align}
In the end, we perform weighted aggregation between the $s_i^{cos}$ from PCB and $s_i$ from few-shot detector for final classification score $s_i^{\ddagger}$ as follow:
\begin{align}
s_i^{\ddagger} = \alpha \cdot s_i + (1-\alpha) \cdot s_i^{cos}  \label{eq:final_score}
\end{align}
where $\alpha$ is the trade-off hyper-parameter.

Moreover, we do not share any parameters between the few-shot detector and PCB module, so that the PCB can not only preserve the quality of classification-aimed translation invariance feature, but also better decouple the classification task and regression task within the RCNN. Furthermore, since the PCB module is offline without any further training, it can be plug-and-play and easily equipped to any other architectures to build stronger few-shot detectors.

%-------------------------------------------------------------------------
\section{Experiments}
In this section, we first introduce the experimental settings in Sec.\ref{sec: setting} and then compare our approach with previous SOTAs on multiple benchmarks in Sec.\ref{sec: comparsion}. Finally, we provide comprehensive ablation studies in Sec.\ref{sec: ablation}.

\subsection{Experimental Setting} \label{sec: setting}
\minisection{Existing benchmarks.}
We follow the previous work \cite{kang2019few, wang2020frustratingly, Xiao2020FSDetView} and utilize the same data splits with \cite{wang2020frustratingly} to evaluate our approach for a fair comparison.
As for PASCAL VOC, we have three random split groups and each of them covers 20 categories, which are randomly divided into 15 base classes and 5 novel classes. Each novel category has $K=1, 2, 3, 5, 10$ objects sampled from the combination of VOC07 and VOC12 train/val set for few-shot training.  And VOC07 test set is used for evaluation. As for COCO, the 60 categories disjoint with VOC are denoted as base classes while the remaining 20 classes are used as novel classes with 
$K=1, 2, 3, 5, 10, 30$ shots. We utilize 5k images from the validation set for evaluation and the rest for training. 

\minisection{Evaluation setting.}
We take two popular evaluation protocols into consideration to access the effectiveness of our approach, including few-shot object detection (\textit{\textbf{FSOD}}) and generalized few-shot object detection (\textit{\textbf{G-FSOD}}). The former protocol is widely adopted by most previous methods \cite{chen2018lstd, kang2019few, Xiao2020FSDetView, yan2019meta} and only focuses on the performance of novel classes. Yet, the latter presents to not only observe the performance on novel classes, but also base and overall performance of the few-shot detector, which is more comprehensive and monitors the occurrence of catastrophic forgetting \cite{wang2020frustratingly}. For evaluation metrics, we report $AP_{50}$ for VOC and  the COCO-style $mAP$ for COCO. Moreover, all results are averaged over multiple repeated runs.

\minisection{Implementation details.}
Our approach employs Faster R-CNN \cite{ren2015faster} (termed as FRCN) as the  basic detection framework and ResNet-101 \cite{he2016deep} pre-trained on ImageNet \cite{russakovsky2015imagenet} as the backbone. We adopt SGD to optimize our network end-to-end with a mini-batch size of 16, momentum of 0.9 and weight decay of $5e^{-5}$.  The learning rate is set to 0.02 during base training and 0.01 during few-shot fine-tuning.  Moreover, the $\lambda$ in GDL of RPN is set to 0 for stopping gradient and  the $\lambda$ in GDL of RCNN is set to 0.75 during base training and 0.01 during novel fine-tuning for scaling gradient. The $\alpha$ in PCB is uniformly set to 0.5 in all settings. 

%--------------------------------------------------------------------------
\begin{table*}[] \centering
	%\label{tab:1}
	%\resizebox{500pt}{107pt}
	\setlength{\tabcolsep}{1.43mm}
	%\resizebox{450pt}{90pt}
	\scalebox{0.94}
	{\begin{tabular}{c|l|l|l|c|ccccc|ccccc|ccccc}
			%\hline
			\toprule[1.1pt]
			\multicolumn{4}{c|}{}            & \multicolumn{1}{c|}{}                    & \multicolumn{5}{c|}{Novel Set 1}                                              & \multicolumn{5}{c|}{Novel Set 2}                                              & \multicolumn{5}{c}{Novel Set 3}                                              \\
			\multicolumn{4}{c|}{\multirow{-2}{*}{Method / Shots}} &\multicolumn{1}{c|}{\multirow{-2}{*}{\textit{w/G}}} & 1             & 2             & 3             & 5             & 10            & 1             & 2             & 3             & 5             & 10            & 1             & 2             & 3             & 5             & 10            \\ \midrule[0.9pt]
			\multicolumn{4}{c|}{YOLO-ft \cite{kang2019few}}      & \xmark     & 6.6           & 10.7          & 12.5          & 24.8          & 38.6          & 12.5          & 4.2           & 11.6          & 16.1          & 33.9          & 13.0          & 15.9          & 15.0          & 32.2          & 38.4          \\
			\multicolumn{4}{c|}{FRCN-ft \cite{yan2019meta}}      & \xmark     & 13.8          & 19.6          & 32.8          & 41.5          & 45.6          & 7.9           & 15.3          & 26.2          & 31.6          & 39.1          & 9.8           & 11.3          & 19.1          & 35.0          & 45.1          \\
			\multicolumn{4}{c|}{LSTD \cite{chen2018lstd}}        & \xmark           & 8.2           & 1.0           & 12.4          & 29.1          & 38.5          & 11.4          & 3.8           & 5.0           & 15.7          & 31.0          & 12.6          & 8.5           & 15.0          & 27.3          & 36.3          \\
			\multicolumn{4}{c|}{FSRW \cite{kang2019few}}         & \xmark          & 14.8          & 15.5          & 26.7          & 33.9          & 47.2          & 15.7          & 15.2          & 22.7          & 30.1          & 40.5          & 21.3          & 25.6          & 28.4          & 42.8          & 45.9          \\
			\multicolumn{4}{c|}{MetaDet \cite{wang2019meta}}     & \xmark           & 18.9          & 20.6          & 30.2          & 36.8          & 49.6          & {21.8}          & 23.1          & 27.8          & 31.7          & 43.0          & 20.6          & 23.9          & 29.4          & 43.9          & 44.1          \\
			\multicolumn{4}{c|}{Meta R-CNN \cite{yan2019meta}}   & \xmark          & 19.9          & 25.5          & 35.0          & 45.7          & 51.5          & 10.4          & 19.4          & 29.6          & 34.8          & 45.4          & 14.3          & 18.2          & 27.5          & 41.2          & 48.1          \\
			\multicolumn{4}{c|}{TFA  \cite{wang2020frustratingly}} & \xmark             & 39.8          & \textbf{\color{blue}36.1}          & 44.7          & \textbf{\color{blue}55.7}          & 56.0          & 23.5          & \textbf{\color{blue}26.9}          & 34.1          & 35.1          & 39.1          & 30.8          & \textbf{\color{blue}34.8}          & \textbf{\color{blue}42.8}          & \textbf{\color{blue}49.5}          & \textbf{\color{blue}49.8}          \\
			\multicolumn{4}{c|}{MPSR  \cite{wu2020multi}} & \xmark             & \textbf{\color{blue}41.7}          & -          & \textbf{\color{blue}51.4}          & 55.2          & \textbf{\color{red}61.8}          & \textbf{\color{blue}24.4}          &-          & \textbf{\color{blue}39.2}          & \textbf{\color{blue}39.9}          & \textbf{\color{blue}47.8}          & \textbf{\color{blue}35.6}          & -         & 42.3          & 48.0          & 49.7         \\
			\midrule[0.9pt]
			\rowcolor[HTML]{EFEFEF}
			\multicolumn{4}{c|}{\cellcolor[HTML]{EFEFEF}DeFRCN (Ours)} & \xmark   & {\cellcolor[HTML]{EFEFEF}\textbf{\color{red} 53.6}} & \textbf{\color{red} 57.5} & \textbf{\color{red} 61.5} & \textbf{\color{red} 64.1} & \textbf{\color{blue}60.8} & \textbf{\color{red} 30.1} & \textbf{\color{red} 38.1} & \textbf{\color{red} 47.0} & \textbf{\color{red} 53.3} & \textbf{\color{red} 47.9} & \textbf{\color{red} 48.4} & \textbf{\color{red} 50.9} & \textbf{\color{red} 52.3} & \textbf{\color{red} 54.9} & \textbf{\color{red}57.4} \\ 
			\midrule[0.9pt]
			% \midrule[0.9pt]
			\multicolumn{4}{c|}{FRCN-ft \cite{yan2019meta}}      & \cmark     & 9.9         & 15.6          & 21.6          & 28.0          & 52.0          & 9.4           & 13.8          & 17.4          & 21.9          & 39.7          & 8.1           & 13.9          & 19.0          & 23.9          & 44.6          \\
			\multicolumn{4}{c|}{FSRW \cite{kang2019few}}         & \cmark          & 14.2          & 23.6         & 29.8          & 36.5          & 35.6          & 12.3          & 19.6          & 25.1          & 31.4          & 29.8          & 12.5          & 21.3          & 26.8          & 33.8          & 31.0          \\
			
			\multicolumn{4}{c|}{TFA \cite{wang2020frustratingly}}     & \cmark         & \textbf{\color{blue}25.3}          & \textbf{\color{blue}36.4}          & 42.1          & 47.9          & 52.8          & 18.3          & \textbf{\color{blue}27.5}          & 30.9          & 34.1          & 39.5          & 17.9          & 27.2          & 34.3          & 40.8          & 45.6          \\
			\multicolumn{4}{c|}{FSDetView \cite{Xiao2020FSDetView}}        & \cmark      & 24.2          & 35.3          &  \textbf{\color{blue}42.2}          & \textbf{\color{blue}49.1}          & \textbf{\color{blue}57.4}          & \textbf{\color{blue}21.6}          & 24.6          & \textbf{\color{blue}31.9}          &\textbf{\color{blue}37.0}          & \textbf{\color{blue}45.7}         & \textbf{\color{blue}21.2}          & \textbf{\color{blue}30.0}          & \textbf{\color{blue}37.2}          & \textbf{\color{blue}43.8}          & \textbf{\color{blue}49.6}  \\
			\midrule[0.9pt]
			\rowcolor[HTML]{EFEFEF}
			\multicolumn{4}{c|}{\cellcolor[HTML]{EFEFEF}DeFRCN (Ours)}   & \cmark  & {\cellcolor[HTML]{EFEFEF}\textbf{\color{red} 40.2}} & \textbf{\color{red} 53.6} & \textbf{\color{red} 58.2} & \textbf{\color{red} 63.6} & \textbf{\color{red} 66.5} & \textbf{\color{red} 29.5} & \textbf{\color{red} 39.7} & \textbf{\color{red} 43.4} & \textbf{\color{red} 48.1} & \textbf{\color{red} 52.8} & \textbf{\color{red} 35.0} & \textbf{\color{red} 38.3} & \textbf{\color{red} 52.9} & \textbf{\color{red} 57.7} & \textbf{\color{red} 60.8} \\ 
			\bottomrule[1.1pt]
	\end{tabular}}
	\vspace{-0.18cm}
	\caption{Experimental results on  VOC dataset. We evaluate DeFRCN performance ($AP_{50}$) on three different splits. The term \textit{w/G} indicates whether we use the \textit{G-FSOD} setting \cite{wang2020frustratingly}. \textbf{\color{red} RED}\textbf{/}\textbf{\color{blue} BLUE} indicate SOTA/the second best. 
		Note that our results are averaged over multiple runs and the base/overall performance are presented in supplementary materials, the same below.
	}
	\vspace{-0.35cm}
	\label{tab:voc-result}
\end{table*}

\begin{table}[] \centering
	\setlength{\tabcolsep}{1.3mm}
	\scalebox{0.91}
	{\begin{tabular}{c|l|l|l|c|cccccc}
			%\hline
			\toprule[1.1pt]
			\multicolumn{4}{c|}{}            & \multicolumn{1}{c|}{}                    & \multicolumn{6}{c}{Shot Number}                                       \\
			\multicolumn{4}{c|}{\multirow{-2}{*}{Method / Shots}} &\multicolumn{1}{c|}{\multirow{-2}{*}{\textit{w/G}}} & 1        & 2              & 3             & 5             & 10    &30                \\ \midrule[0.9pt]
			\multicolumn{4}{c|}{FRCN-ft \cite{yan2019meta}}      & \xmark     & 1.0$^\ast$     & 1.8$^\ast$     & 2.8$^\ast$          & 4.0$^\ast$          & 6.5          &11.1            \\
			\multicolumn{4}{c|}{FSRW \cite{kang2019few}}         & \xmark          & -    & -      & -          & -          & 5.6          & 9.1                  \\
			\multicolumn{4}{c|}{MetaDet \cite{wang2019meta}}     & \xmark           & -     & -     & -          & -          & 7.1          & 11.3                  \\
			\multicolumn{4}{c|}{Meta R-CNN \cite{yan2019meta}}   & \xmark          & -     & -     & -          & -          &  8.7         & 12.4                  \\
			\multicolumn{4}{c|}{TFA  \cite{wang2020frustratingly}} & \xmark             & 4.4$^\ast$  &5.4$^\ast$       &6.0$^\ast$          & 7.7$^\ast$          &10.0         & 13.7                 \\
			\multicolumn{4}{c|}{MPSR  \cite{wu2020multi}} & \xmark             & \textbf{\color{blue}5.1}$^\ast$  &\textbf{\color{blue}6.7}$^\ast$       & \textbf{\color{blue}7.4}$^\ast$          & 8.7$^\ast$          &9.8         & 14.1                 \\
			\multicolumn{4}{c|}{FSDetView \cite{Xiao2020FSDetView}}        & \xmark      &4.5  &6.6       &7.2          & \textbf{\color{blue}10.7}           & \textbf{\color{blue}12.5}         & \textbf{\color{blue}14.7}           \\
			\midrule[0.9pt]
			\rowcolor[HTML]{EFEFEF}
			\multicolumn{4}{c|}{\cellcolor[HTML]{EFEFEF}DeFRCN (Ours)} & \xmark   & {\cellcolor[HTML]{EFEFEF}\textbf{\color{red} 9.3}} & \textbf{\color{red} 12.9} & \textbf{\color{red} 14.8} & \textbf{\color{red} 16.1} & \textbf{\color{red} 18.5} & \textbf{\color{red} 22.6} \\ 
			\midrule[0.9pt]
			% \midrule[0.9pt]
			\multicolumn{4}{c|}{FRCN-ft \cite{yan2019meta}}      & \cmark     & 1.7         & 3.1  & 3.7          & 4.6          & 5.5          & 7.4                  \\
			\multicolumn{4}{c|}{TFA \cite{wang2020frustratingly}}     & \cmark         & 1.9         & 3.9  & 5.1          & 7.0          & 9.1          & 12.1                 \\
			\multicolumn{4}{c|}{FSDetView \cite{Xiao2020FSDetView}}        & \cmark      & \textbf{\color{blue}3.2}         & \textbf{\color{blue}4.9}   & \textbf{\color{blue}6.7}          &  \textbf{\color{blue}8.1}          & \textbf{\color{blue}10.7}         & \textbf{\color{blue}15.9}           \\
			\midrule[0.9pt]
			\rowcolor[HTML]{EFEFEF}
			\multicolumn{4}{c|}{\cellcolor[HTML]{EFEFEF}DeFRCN (Ours)}   & \cmark  & {\cellcolor[HTML]{EFEFEF}\textbf{\color{red} 4.8}} & \textbf{\color{red} 8.5} & \textbf{\color{red} 10.7} & \textbf{\color{red} 13.6} & \textbf{\color{red} 16.8} & \textbf{\color{red} 21.2}  \\ 
			\bottomrule[1.1pt]
	\end{tabular}}
	\vspace{-0.18cm}
	\caption{Experimental results on COCO dataset. We evaluate DeFRCN performance ($mAP$) over multiple runs. The superscript $^\ast$ indicates that the results are reproduced by us. 
	}
	\vspace{-0.54cm}
	\label{tab:coco-result}
\end{table}
%--------------------------------------------------------------------------

\subsection{Comparison Results} \label{sec: comparsion}

\minisection{PASCAL VOC.}
We present our evaluation results of VOC on three different data splits in Table \ref{tab:voc-result}. It can be seen that, no matter under the \textit{FSOD} or \textit{G-FSOD} setting, our DeFRCN is significantly superior to the recent state-of-the-art approaches by a large margin (up to 21.4$\%$), which demonstrates the effectiveness of our approach. Based on the results of Table \ref{tab:voc-result}, we further notice that two interesting phenomena exist in few-shot detection: (1) For \textit{FSOD} setting, the increment of novel shots does not necessarily lead to an advance in final performance. Take Novel Set 1 as an example, the $AP_{50}$ of 5-shot is 64.1$\%$ but 10-shot is 60.8$\%$ (-3.3$\%$). There is a similar case in TFA. We conjecture that the quality of sample is vital in data-scarce scenario and adding low-quality samples may be harmful to the detector. (2) For the comparison between \textit{FSOD} and \textit{G-FSOD}, we find that as the number of shots increases, the final performance of \textit{G-FSOD} grows faster than that of \textit{FSOD} (40.2$\%$ $\to$ 66.5$\%$  \textit{vs.} 53.6$\%$ $\to$ 60.8$\%$), which is due to the addition of more negative samples under the \textit{G-FSOD} setting.

%--------------------------------------------------------------------------
\begin{table}[] \centering
	\setlength{\tabcolsep}{1.2mm}
	\scalebox{0.87}
	{\begin{tabular}{c|ccccc|c}
			%\hline
			\toprule[1.1pt]
			\multicolumn{1}{c|}{\multirow{-1}{*}{Method}}  & FRCN-ft  & FSRW       & MetaDet            & MetaRCNN             & MPSR             & Ours                 \\ \midrule[0.9pt]
			\multicolumn{1}{c|}{$mAP$}           & 31.2  & 32.3    & 33.9      & 37.4          & 42.3          & \cellcolor[HTML]{EFEFEF}{\textbf{55.9}}                            \\
			\bottomrule[1.1pt]
	\end{tabular}}
	\vspace{-0.18cm}
	\caption{
		The 10-shot cross-domain \textit{FSOD} performance on COCO base set $\to$ VOC novel set. All detection results for comparison refer from \cite{kang2019few,wu2020multi,  yan2019meta}.
	}
	\vspace{-0.6cm}
	\label{tab:coco-to-voc}
\end{table}

\minisection{COCO.}
The Table \ref{tab:coco-result} shows all evaluation results on COCO dataset with the standard COCO-style averaged AP ($mAP$). Obviously, our approach consistently outperforms recent SOTAs in all setups, including \textit{FSOD} and \textit{G-FSOD} for $K$=1,2,3,5,10,30. For \textit{FSOD}, we achieve around 6.0$\%$ and 7.9$\%$ improvement over the best method in 10-shot and 30-shot respectively, which demonstrates the strong robustness and generalization ability of our method in the few-shot scenario. Furthermore, compared to the fine-tuning based methods, the number of learnable parameters of DeFRCN is almost the same as FRCN-ft and much more than TFA. The results in Table \ref{tab:coco-result} reveal that our method not only guarantees the sufficient learning of these parameters, but also does not fall into the severe over-fitting. All base/overall results of \textit{G-FSOD} are presented in supplementary materials.

\minisection{COCO to VOC.}
We conduct the cross-domain \textit{FSOD} experiments on the standard VOC 2007 test set with following the same setting from \cite{kang2019few, wu2020multi}, which uses the base dataset with 60 classes as in the previous COCO within-domain setting and the novel dataset with 10-shot objects for each of the 20 classes from VOC. As shown in the Table \ref{tab:coco-to-voc}, our approach achieves the best performance with 55.9$\%$, which has 13.6$\%$ improvement than MPSR \cite{wu2020multi}. This huge upswing demonstrates that our proposed DeFRCN has better generalization ability in cross-domain situations.

%--------------------------------------------------------------------------
\begin{table}[] \centering
	\setlength{\tabcolsep}{1.7mm}
	\scalebox{0.86}
	{\begin{tabular}{c|ccc|c|cc}
			%\hline
			\toprule[1.1pt]
			\multicolumn{1}{c|}{}            & \multicolumn{3}{c|}{}                    & \multicolumn{1}{c|}{}         & \multicolumn{2}{c}{Novel}                               \\
			\multicolumn{1}{c|}{\multirow{-2}{*}{FRCN}}      &\multicolumn{1}{c}{\multirow{-2}{*}{GDL-B}} & \multicolumn{1}{c}{\multirow{-2}{*}{GDL-N}}        & \multicolumn{1}{c|}{\multirow{-2}{*}{PCB}}              & \multicolumn{1}{c|}{\multirow{-2}{*}{Base}}              & \multicolumn{1}{c}{10}             &  \multicolumn{1}{c}{30}                 \\
			\midrule[0.9pt]
			\cmark  & &&&&7.9&12.2\\
			\cmark  & &&\cmark&&10.4&14.8 \\
			\cmark  & &\cmark&&&15.2&19.0 \\
			\cmark  & &\cmark&\cmark&\multirow{-4}{*}{38.4}&16.6&20.5 \\
			\midrule[0.9pt]
			\cmark  &\cmark&&&&8.2&13.1 \\
			\cmark  &\cmark&&\cmark&&10.8&15.1 \\
			\cmark  &\cmark&\cmark&&&16.9&21.0 \\
			\rowcolor[HTML]{EFEFEF}
			\cmark  &\cmark&\cmark&\cmark&\multirow{-4}{*}{39.0}&18.5&22.6 \\
			\bottomrule[1.1pt]
	\end{tabular}}
	\vspace{-0.18cm}
	\caption{Effectiveness of different modules in DeFRCN. All results are conducted on COCO dataset. The GDL-B and GDL-N indicates that we use GDL in base training phase and novel fine-tuning phase respectively.}
	\vspace{-0.3cm}
	\label{tab:ablation-study}
\end{table}
%--------------------------------------------------------------------------

%--------------------------------------------------------------------------
\vspace{0.3cm}
\begin{table}[] \centering
	\setlength{\tabcolsep}{1.3mm}
	\scalebox{0.85}
	{\begin{tabular}{c|l|l|l|ccc|ccc}
			\toprule[1.1pt]
			\multicolumn{4}{c|}{Backbone} & $AP$        & $AP_{50}$              & $AP_{75}$         & $AP_{s}$            & $AP_{m}$    &$AP_{l}$                \\ \midrule[0.9pt]
			\multicolumn{4}{c|}{R50-C4-1x \cite{wu2019detectron2}}       & 35.7     & 56.1     & 38.0          & 19.2         & 40.9          &48.7            \\
			\rowcolor[HTML]{EFEFEF}
			\multicolumn{4}{c|}{+ GDL}            & 36.5     & \textbf{57.6}     & 39.2          & 19.8          & 41.7          & 50.3           \\
			\midrule[0.9pt]
			\multicolumn{4}{c|}{R101-C4-3x \cite{wu2019detectron2}}       & 41.1     & 61.4     & 44.0          & 22.2          & 45.5          &55.9            \\
			\rowcolor[HTML]{EFEFEF}
			\multicolumn{4}{c|}{+ GDL}            & 41.9     & \textbf{62.3}     & 45.1          & 22.3          & 46.6          & 57.8           \\
			\bottomrule[1.1pt]
	\end{tabular}}
	\vspace{-0.18cm}
	\caption{Conventional object detection results on COCO. }
	\vspace{-0.4cm}
	\label{tab:coco-3x}
\end{table}

\subsection{Ablation Study} \label{sec: ablation}

\minisection{Effectiveness of different modules.}
We conduct relative ablations in 10/30-shot scenarios on the COCO dataset to carefully analyze how much each module contributes to the ultimate performance of DeFRCN. All results are shown in Table \ref{tab:ablation-study} in great details. 
Specifically, the first row shows the results of plain FRCN, which only achieves  7.9$\%$/12.2$\%$ for 10/30-shot respectively, indicating that the original model without any few-shot techniques is severely over-fitting due to the lack of training data. Next, we take four progressive steps to complete the exploration of our DeFRCN: (1) add GDL in base training phase (GDL-B). Through the results of rows 1-4 and 5-8, we find that the GDL-B improves by 0.6$\%$ on base classes and also a certain improvement (0.3$\%$ $\sim$ 2.1$\%$) on novel classes. This indicates that a better base model is beneficial to the performance of few-shot detector. (2) add GDL in novel fine-tuning phase (GDL-N). The results of first row and third row show that GDL-N makes an amazing boost with 7.3$\%$/6.8$\%$ for 10/30-shot, which are mainly from two aspects: i) more learnable parameters guarantee sufficient ability to transfer to novel domain, and ii) GDL greatly reduces the risk of over-fitting.(3) add PCB in the inference phase. As a plug-and-play module, PCB is orthogonal to GDL, so no matter which setting PCB is added, our model further gains 1.4$\%$ $\sim$ 2.6$\%$ points on $mAP$. (4) Finally, we integrate the above three modules into original FRCN, and the last line shows the final performance of DeFRCN. Compared to the plain results in the first row, we obtain a marvelous promotion of \textbf{10.6$\%$/10.4$\%$} for 10/30-shot, which proves the effectiveness of our approach.

%--------------------------------------------------------------------------
\begin{figure}[]
	\centering
	% 	\vspace{-0.2cm}
	\hspace{-0.3cm}
	\subfigure[The base training stage]{ 
		\begin{minipage}{0.45\linewidth}  \label{fig:lambda-base}
			\centering
			\includegraphics[scale=0.31]{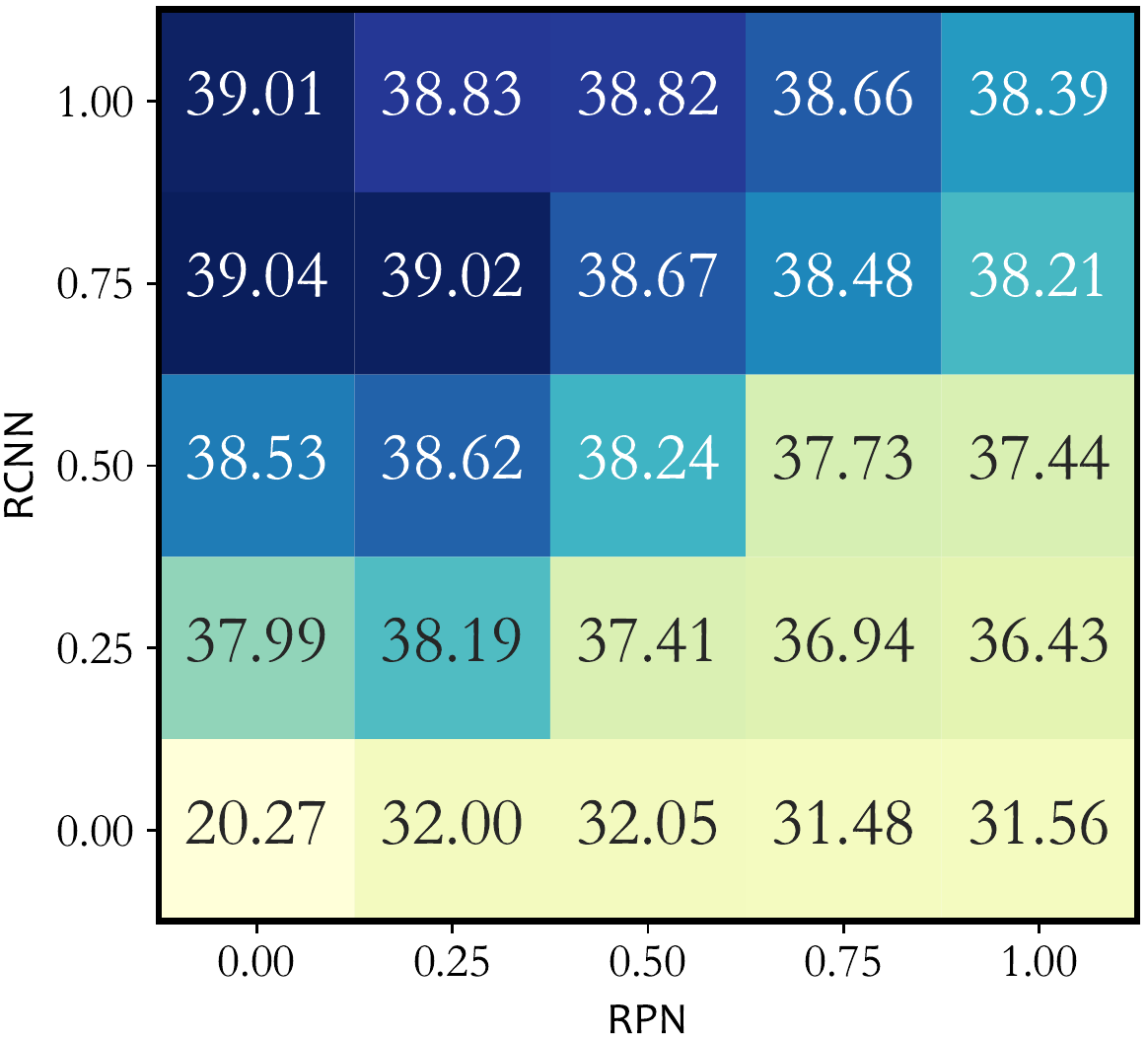} 
	    \end{minipage}}
	\subfigure[The novel fine-tune stage]{  
		\begin{minipage}{0.45\linewidth}  \label{fig:lambda-novel}
			\centering       
			\includegraphics[scale=0.31]{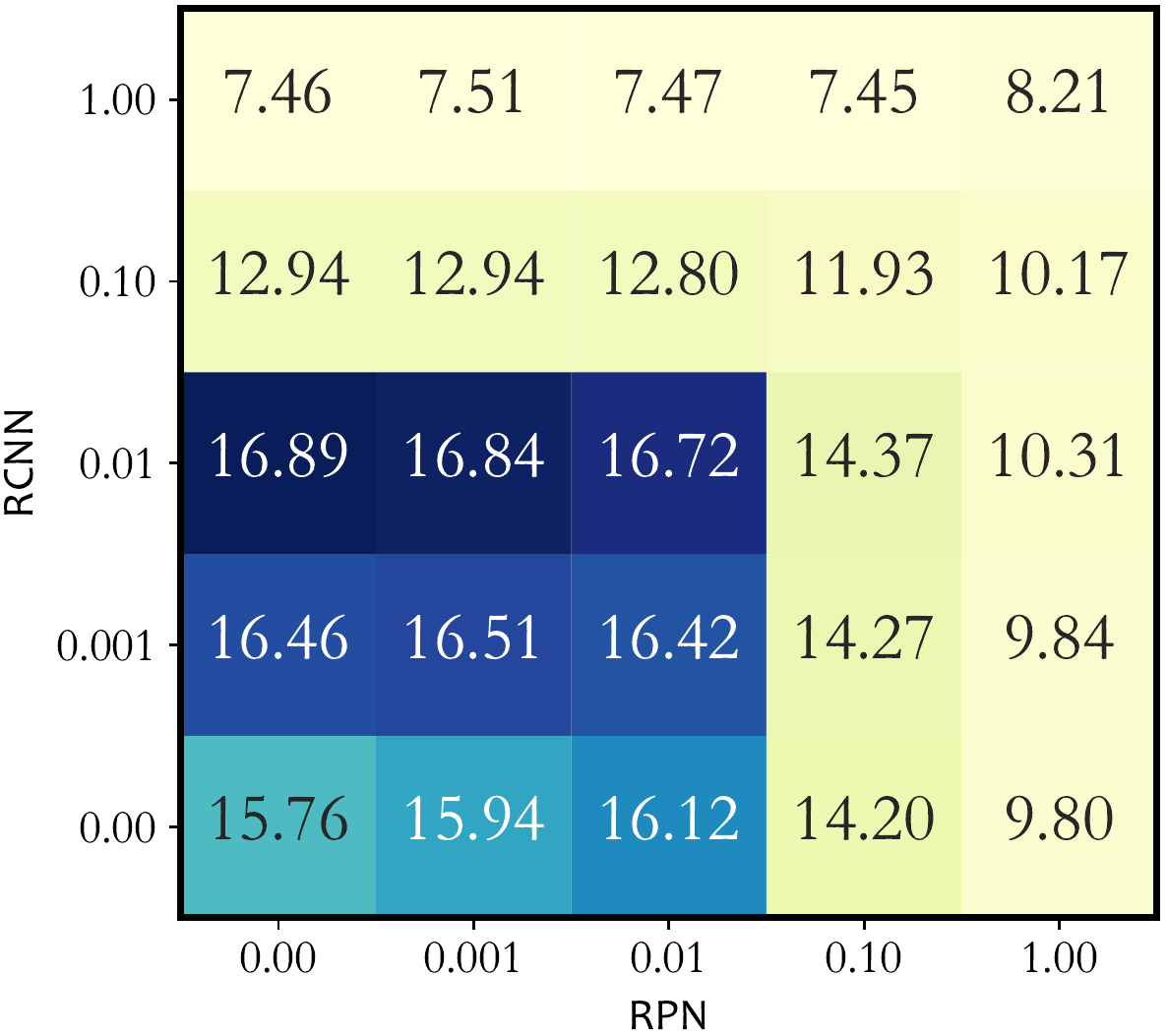}
	    \end{minipage}}
	\vspace{-0.3cm}
	\caption{The effectiveness of different degree of coupling. The horizontal and vertical axis represent the $\lambda$ in GDL of RPN and RCNN respectively. Note the results in (b) do not use PCB to ensure the impact of GDL is considered only.}
	\label{fig:lambda} 
	\vspace{-0.5cm}
\end{figure}
%--------------------------------------------------------------------------

\minisection{Effectiveness of the degree of decoupling.}
We carefully explore the influence of decoupling with setting different $\lambda_{rpn}$ and $\lambda_{rcnn}$ in GDL during both base training and novel fine-tuning, and all results are illustrated in Fig.\ref{fig:lambda}. No matter in the base training or the novel fine-tuning stage, the model tends to achieve higher performance when $\lambda_{rpn}$ is set to a smaller value (close to 0), while $\lambda_{rcnn}$ needs an appropriate value to ensure that the backbone can be optimized better. This observation prompts us to perform \textbf{\textit{stop-gradient}} for RPN and \textbf{\textit{scale-gradient}} for RCNN in DeFRCN. In addition, we further get a very interesting conclusion from four corners in Fig.\ref{fig:lambda-base}: in term of FRCN backbone optimization, RPN plays a negative role in this procedure (39.01 \textit{vs.} 38.39), while RCNN has a positive effect (31.56 \textit{vs.} 38.39). 

\minisection{Can GDL boost conventional detection?}
The above analysis shows that GDL brings a significant improvement on FSOD. Since the problem it solves (that is, the contradiction in Faster R-CNN) also potentially exists in conventional detection, we conjecture that our GDL is as well as effective in data-sufficient scenarios. Thus we further conduct experiments on COCO 2017 dataset with standard setup \cite{ren2015faster}  and the results are shown in Table \ref{tab:coco-3x}. It can be seen that the proposed GDL outperforms baselines on all evaluation metrics. Specifically, adding GDL to original FRCN gains 1.5$\%$ and 0.9$\%$ AP50 for Res-50 and Res-101 respectively.

\section{Conclusion}
In this paper, we look closely into the visual task of few-shot object detection and propose a simple yet effective fine-tuning based framework, named Decoupled Faster R-CNN, which remarkably alleviates the potential contradictions of conventional Faster R-CNN in data-scarce scenario with introducing novel GDL and PCB. Despite its simplicity, our method still achieves new state-of-the-art on various benchmarks, which demonstrates its effectiveness and versatility.

\vspace{0.2cm}
\noindent
\textbf{Acknowledgement. } This paper is supported by the National Key R\&D Plan of the Ministry of Science and Technology (Project No. 2020AAA0104400).

\clearpage
\appendix
\section*{\centering Supplementary Material}
In this supplementary material, we provide additional details which we could not include in the main paper due to space limitations, including more experimental analysis and visualization details that help us develop further insights to the proposed approach. We discuss:

\begin{itemize}
	\setlength\itemsep{-0.15em}
	\item More results of generalized few-shot object detection.
	\item Additional analysis on Prototypical Calibration Block.
	\item Related extensions of Gradient Decoupled Layer.
	\item Qualitative visualization results of our approach.
\end{itemize}

\section{Generalized Few-Shot Object Detection}
\subsection{Implementation Details}
As mentioned in the main paper, we take two popular evaluation protocols into consideration to assess the effectiveness of our approach, including few-shot object detection (\textit{\textbf{FSOD}}) and generalized few-shot object detection (\textit{\textbf{G-FSOD}}). The difference between these two protocols is whether the performance of base classes is still required after the fine-tuning stage. Following the \textit{G-FSOD} setting in TFA \cite{wang2020frustratingly}, we fine-tune our DeFRCN on  a small balanced training set consisting of both base and novel classes, where each class has the same number of annotated objects ($\ie$, $K$-shot). In addition to deploying more training iterations ($2\times$), other experimental settings for \textit{G-FSOD} are exactly consistent with the \textit{FSOD} in our paper.

\subsection{Experimental Results of \textit{\textbf{G-FSOD}} Setting}
In this section, we show the full benchmark results of the \textit{G-FSOD} setting. For each evaluation metric, we report the average results of $n$ random splits ($n=30$ for VOC and $n=10$ for  COCO) with the same data split in TFA as well as the 95\% confidence interval estimate of the mean values.

\minisection{PASCAL VOC.}
We present the complete \textit{G-FSOD} results of VOC ($K=1,2,3,5,10$) in Table \ref{tab:gfsod-voc} and then analyze our results from the following three aspects: 
(1) Novel AP. The novel AP of our model is usually over 7$\%$ points higher than that of TFA in three data splits, which indicates that the proposed DeFRCN has absolute advantage on novel performance. 
(2) Base AP. Our approach  is able to outperform TFA on split 2 (+1.9$\%$ $\sim$ +3.7$\%$ $AP$), however, it is slightly worse on data split 1 and 3 (-0.3$\%$  $\sim$  -1.0$\%$ $AP$). We notice that the base performance advantage of TFA comes from the strategy of fine-tuning only the last layer of detectors, which can indeed be eccentric to ensure that the base performance does not decrease too much, but it also results in the novel performance cannot be further improved. 
(3) Overall AP. As shown in the Table \ref{tab:gfsod-voc}, the proposed DeFRCN achieves the best overall performance across all settings (+1.4$\%$ $\sim$ +4.0$\%$ $AP$), including data splits and shots.

%--------------------------------------------------------------------------
% VOC G-FSOD Result
\begin{table*}[htb] \centering
	\setlength{\tabcolsep}{4mm}
	\scalebox{0.78}{
		\begin{tabular}{c|c|c|ccc|c|c}
			\toprule
			\multirow{2}{*}{Split} & \multirow{2}{*}{\# shots} &\multirow{2}{*}{Method} &  \multicolumn{3}{c|}{Overall \textit{\#20}}  &\multicolumn{1}{c|}{Base \textit{\#15}} & \multicolumn{1}{c}{Novel \textit{\#5}} \\ \cmidrule{4-8}
			& & & AP & AP50 & AP75 & AP & AP \\ \midrule
			\multirow{21}{*}{Split 1} & 
			\multirow{4}{*}{1}
			& FSRW~\cite{kang2019few} & 27.6$ \pm $0.5 & 50.8 $\pm $0.9 & 26.5$ \pm $0.6 & 34.1$ \pm $0.5 & 8.0$ \pm $1.0 \\
			& & FRCN+ft \cite{yan2019meta}& 30.2$\pm$0.6 & 49.4$\pm$0.7 & 32.2$\pm$0.9 & 38.2$\pm$0.8 & 6.0$\pm$0.7 \\
			& & TFA \cite{wang2020frustratingly} & 40.6$\pm$0.5 & 64.5$\pm$0.6 & 44.7$\pm$0.6 & \textbf{49.4$\pm$0.4} & 14.2$\pm$1.4  \\ 
			& &\cellcolor{Gray} {DeFRCN} & \cellcolor{Gray}\textbf{42.0$\pm$0.6} \textbf{\scriptsize{\color{red}(+1.4)}} & \cellcolor{Gray} \textbf{66.7$\pm$0.8} \textbf{\scriptsize{\color{red}(+2.2)}} & \cellcolor{Gray} \textbf{45.5$\pm$0.7}  \textbf{\scriptsize{\color{red}(+0.8)}} & \cellcolor{Gray} 48.4$\pm$0.4  \textbf{\scriptsize{\color{blue}(-1.0)}} & \cellcolor{Gray} \textbf{22.5$\pm$1.7} \textbf{\scriptsize{\color{red}(+8.3)}} \\
			\cmidrule{2-8}
			& \multirow{4}{*}{2} & FSRW~\cite{kang2019few} & 
			28.7$\pm$0.4&52.2$\pm$0.6&27.7$\pm$0.5&33.9$\pm$0.4&13.2$\pm$1.0 \\
			& & FRCN+ft \cite{yan2019meta} & 30.5$\pm$0.6 & 49.4$\pm$0.8 & 32.6$\pm$0.7 & 37.3$\pm$0.7 & 9.9$\pm$0.9  \\
			& &TFA \cite{wang2020frustratingly} & 42.6$\pm$0.3 & 67.1$\pm$0.4 &47.0$\pm$0.4 & \textbf{49.6$\pm$0.3} &21.7$\pm$1.0 \\ 
			& &\cellcolor{Gray} {DeFRCN} & \cellcolor{Gray}\textbf{44.3$\pm$0.4} \textbf{\scriptsize{\color{red}(+1.7)}} & \cellcolor{Gray} \textbf{70.2$\pm$0.5} \textbf{\scriptsize{\color{red}(+3.1)}}& \cellcolor{Gray} \textbf{48.0$\pm$0.6} \textbf{\scriptsize{\color{red}(+1.0)}}& \cellcolor{Gray} 49.1$\pm$0.3 \textbf{\scriptsize{\color{blue}(-0.5)}}& \cellcolor{Gray} \textbf{30.6$\pm$1.2} \textbf{\scriptsize{\color{red}(+8.9)}} \\
			\cmidrule{2-8}
			& \multirow{4}{*}{3} & FSRW~\cite{kang2019few} & 29.5$\pm$0.3&53.3$\pm$0.6&28.6$\pm$0.4&33.8$\pm$0.3&16.8$\pm$0.9 \\
			& & FRCN+ft \cite{yan2019meta} & 31.8$\pm$0.5 & 51.4$\pm$0.8 & 34.2$\pm$0.6 & 37.9$\pm$0.5  & 13.7$\pm$1.0  \\
			& &TFA  \cite{wang2020frustratingly} & 43.7$\pm$0.3 & 68.5$\pm$0.4 & 48.3$\pm$0.4 &\textbf{49.8$\pm$0.3} &25.4$\pm$0.9  \\			
			& &\cellcolor{Gray} {DeFRCN} & \cellcolor{Gray}\textbf{45.3$\pm$0.3} \textbf{\scriptsize{\color{red}(+1.6)}} & \cellcolor{Gray} \textbf{71.5$\pm$0.4} \textbf{\scriptsize{\color{red}(+3.0)}} & \cellcolor{Gray} \textbf{49.0$\pm$0.5} \textbf{\scriptsize{\color{red}(+0.7)}} & \cellcolor{Gray} 49.3$\pm$0.3 \textbf{\scriptsize{\color{blue}(-0.5)}} & \cellcolor{Gray} \textbf{33.7$\pm$0.8}  \textbf{\scriptsize{\color{red}(+8.3)}} \\ \cmidrule{2-8}
			& \multirow{4}{*}{5} & FSRW~\cite{kang2019few} & 30.4$\pm$0.3&54.6$\pm$0.5&29.6$\pm$0.4&33.7$\pm$0.3&20.6$\pm$0.8 \\
			& & FRCN+ft \cite{yan2019meta} & 32.7$\pm$0.5 & 52.5$\pm$0.8 & 35.0$\pm$0.6 & 37.6$\pm$0.4  & 17.9$\pm$1.1  \\
			& &TFA  \cite{wang2020frustratingly} & 44.8$\pm$0.3 &70.1$\pm$0.4 & 49.4$\pm$0.4 & \textbf{50.1$\pm$0.2 }& 28.9$\pm$0.8  \\
			& &\cellcolor{Gray} {DeFRCN} & \cellcolor{Gray}\textbf{46.4$\pm$0.3} \textbf{\scriptsize{\color{red}(+1.6)}} & \cellcolor{Gray} \textbf{73.1$\pm$0.3} \textbf{\scriptsize{\color{red}(+3.0)}} & \cellcolor{Gray} \textbf{50.4$\pm$0.4} \textbf{\scriptsize{\color{red}(+1.0)}} & \cellcolor{Gray} 49.6$\pm$0.3 \textbf{\scriptsize{\color{blue}(-0.5)}} & \cellcolor{Gray} \textbf{37.3$\pm$0.8 } \textbf{\scriptsize{\color{red}(+8.4)}} \\ \cmidrule{2-8}
			& \multirow{3}{*}{10} & FRCN+ft \cite{wang2020frustratingly} & 33.3$\pm$0.4 & 53.8$\pm$0.6 & 35.5$\pm$0.4 & 36.8$\pm$0.4 & 22.7$\pm$0.9 \\
			& &TFA  \cite{wang2020frustratingly} & 45.8$\pm$0.2 & 71.3$\pm$0.3 & 50.4$\pm$0.3 &\textbf{50.4$\pm$0.2} & 32.0$\pm$0.6  \\ 
			& &\cellcolor{Gray} {DeFRCN} & \cellcolor{Gray}\textbf{47.2$\pm$0.2} \textbf{\scriptsize{\color{red}(+1.4)}} & \cellcolor{Gray} \textbf{74.0$\pm$0.3} \textbf{\scriptsize{\color{red}(+2.7)}} & \cellcolor{Gray} \textbf{51.3$\pm$0.3} \textbf{\scriptsize{\color{red}(+0.9)}} & \cellcolor{Gray} 49.9$\pm$0.2 \textbf{\scriptsize{\color{blue}(-0.5)}} & \cellcolor{Gray} \textbf{39.8$\pm$0.7} \textbf{\scriptsize{\color{red}(+7.8)}} \\ \midrule
			\multirow{21}{*}{Split 2} & \multirow{4}{*}{1} & FSRW~\cite{kang2019few} &
			28.4$\pm$0.5&51.7$\pm$0.9&27.3$\pm$0.6&35.7$\pm$0.5&6.3$\pm$0.9 \\
			& & FRCN+ft \cite{yan2019meta} & 30.3$\pm$0.5 & 49.7$\pm$0.5 & 32.3$\pm$0.7 & 38.8$\pm$0.6 & 5.0$\pm$0.6  \\
			& &TFA  \cite{wang2020frustratingly} &36.7$\pm$0.6 & 59.9$\pm$0.8 & 39.3$\pm$0.8 & 45.9$\pm$0.7 & 9.0$\pm$1.2  \\
			& &\cellcolor{Gray} {DeFRCN} & \cellcolor{Gray}\textbf{40.7$\pm$0.5} \textbf{\scriptsize{\color{red}(+4.0)}} & \cellcolor{Gray} \textbf{64.8$\pm$0.7} \textbf{\scriptsize{\color{red}(+4.9)}} & \cellcolor{Gray} \textbf{43.8$\pm$0.6} \textbf{\scriptsize{\color{red}(+4.5)}} & \cellcolor{Gray} \textbf{49.6$\pm$0.4} \textbf{\scriptsize{\color{red}(+3.7)}} & \cellcolor{Gray} \textbf{14.6$\pm$1.5} \textbf{\scriptsize{\color{red}(+5.6)}}  \\ \cmidrule{2-8}
			& \multirow{4}{*}{2} & FSRW~\cite{kang2019few} & 
			29.4$\pm$0.3&53.1$\pm$0.6&28.5$\pm$0.4&35.8$\pm$0.4&9.9$\pm$0.7 \\
			& & FRCN+ft \cite{yan2019meta} & 30.7$\pm$0.5 & 49.7$\pm$0.7 & 32.9$\pm$0.6 & 38.4$\pm$0.5 & 7.7$\pm$0.8 \\
			& & TFA  \cite{wang2020frustratingly} & 39.0$\pm$0.4 & 63.0$\pm$0.5 & 42.1$\pm$0.6 &47.3$\pm$0.4 &14.1$\pm$0.9  \\
			& &\cellcolor{Gray} {DeFRCN} & \cellcolor{Gray}\textbf{42.7$\pm$0.3} \textbf{\scriptsize{\color{red}(+3.7)}} & \cellcolor{Gray} \textbf{67.7$\pm$0.5} \textbf{\scriptsize{\color{red}(+4.7)}} & \cellcolor{Gray} \textbf{45.7$\pm$0.5} \textbf{\scriptsize{\color{red}(+3.6)}} & \cellcolor{Gray} \textbf{50.3$\pm$0.2} \textbf{\scriptsize{\color{red}(+3.0)}} & \cellcolor{Gray} \textbf{20.5$\pm$1.0} \textbf{\scriptsize{\color{red}(+6.4)}} \\ \cmidrule{2-8}
			& \multirow{4}{*}{3} & FSRW~\cite{kang2019few} & 29.9$\pm$0.3&53.9$\pm$0.4&29.0$\pm$0.4&35.7$\pm$0.3&12.5$\pm$0.7 \\
			& & FRCN+ft \cite{yan2019meta} & 31.1$\pm$0.3 & 50.1$\pm$0.5 & 33.2$\pm$0.5 & 38.1$\pm$0.4 & 9.8$\pm$0.9  \\
			& & TFA  \cite{wang2020frustratingly} &40.1$\pm$0.3 & 64.5$\pm$0.5 & 43.3$\pm$0.4 & 48.1$\pm$0.3 & 16.0$\pm$0.8  \\
			& &\cellcolor{Gray} {DeFRCN} & \cellcolor{Gray}\textbf{43.5$\pm$0.3} \textbf{\scriptsize{\color{red}(+3.4)}} & \cellcolor{Gray} \textbf{68.9$\pm$0.4} \textbf{\scriptsize{\color{red}(+4.4)}} & \cellcolor{Gray} \textbf{46.6$\pm$0.4} \textbf{\scriptsize{\color{red}(+3.3)}} & \cellcolor{Gray} \textbf{50.6$\pm$0.3} \textbf{\scriptsize{\color{red}(+2.5)}} & \cellcolor{Gray} \textbf{22.9$\pm$1.0} \textbf{\scriptsize{\color{red}(+6.9)}}  \\ \cmidrule{2-8}
			& \multirow{4}{*}{5} & FSRW~\cite{kang2019few} & 30.4$\pm$0.4&54.6$\pm$0.5&29.5$\pm$0.5&35.3$\pm$0.3&15.7$\pm$0.8 \\
			& & FRCN+ft \cite{yan2019meta} & 31.5$\pm$0.3 & 50.8$\pm$0.7 & 33.6$\pm$0.4 & 37.9$\pm$0.4 & 12.4$\pm$0.9 \\
			& & TFA  \cite{wang2020frustratingly} & 40.9$\pm$0.4 &65.7$\pm$0.5 & 44.1$\pm$0.5 & 48.6$\pm$0.4  & 17.8$\pm$0.8  \\
			& &\cellcolor{Gray} {DeFRCN} & \cellcolor{Gray}\textbf{44.6$\pm$0.3} \textbf{\scriptsize{\color{red}(+3.7)}} & \cellcolor{Gray} \textbf{70.2$\pm$0.5} \textbf{\scriptsize{\color{red}(+4.5)}} & \cellcolor{Gray} \textbf{47.8$\pm$0.4} \textbf{\scriptsize{\color{red}(+3.7)}} & \cellcolor{Gray} \textbf{51.0$\pm$0.2} \textbf{\scriptsize{\color{red}(+2.4)}} &  \cellcolor{Gray} \textbf{25.8$\pm$0.9} \textbf{\scriptsize{\color{red}(+8.0)}} \\ \cmidrule{2-8}
			& \multirow{3}{*}{10} & FRCN+ft \cite{wang2020frustratingly} & 32.2$\pm$0.3 & 52.3$\pm$0.4 & 34.1$\pm$0.4 & 37.2$\pm$0.3 & 17.0$\pm$0.8 \\
			& & TFA  \cite{wang2020frustratingly} & 42.3$\pm$0.3 & 67.6$\pm$0.4 & 45.7$\pm$0.3 & 49.4$\pm$0.2 & 20.8$\pm$0.6  \\ 
			& &\cellcolor{Gray} {DeFRCN} & \cellcolor{Gray}\textbf{45.6$\pm$0.2} \textbf{\scriptsize{\color{red}(+3.3)}} & \cellcolor{Gray} \textbf{71.5$\pm$0.3} \textbf{\scriptsize{\color{red}(+3.9)}} & \cellcolor{Gray} \textbf{49.0$\pm$0.3} \textbf{\scriptsize{\color{red}(+3.3)}} & \cellcolor{Gray} \textbf{51.3$\pm$0.2} \textbf{\scriptsize{\color{red}(+1.9)}} & \cellcolor{Gray} \textbf{29.3$\pm$0.7 } \textbf{\scriptsize{\color{red}(+8.5)}} \\ \midrule
			\multirow{21}{*}{Split 3} & \multirow{4}{*}{1} & FSRW~\cite{kang2019few} &
			27.5$\pm$0.6&50.0$\pm$1.0&26.8$\pm$0.7&34.5$\pm$0.7&6.7$\pm$1.0 \\
			& & FRCN+ft \cite{yan2019meta} & 30.8$\pm$0.6 & 49.8$\pm$0.8 & 32.9$\pm$0.8 & 39.6$\pm$0.8 & 4.5$\pm$0.7 \\
			& & TFA  \cite{wang2020frustratingly} & 40.1$\pm$0.3 & 63.5$\pm$0.6 & 43.6$\pm$0.5 & \textbf{50.2$\pm$0.4} &9.6$\pm$1.1   \\ 			
			& &\cellcolor{Gray} {DeFRCN} & \cellcolor{Gray}\textbf{41.6$\pm$0.5} \textbf{\scriptsize{\color{red}(+1.5)}} & \cellcolor{Gray} \textbf{66.0$\pm$0.9} \textbf{\scriptsize{\color{red}(+2.5)}} & \cellcolor{Gray} 44.9$\pm$0.6 \textbf{\scriptsize{\color{red}(+1.3)}} & \cellcolor{Gray} 49.4$\pm$0.4\textbf{\scriptsize{\color{blue}(-0.8)}} & \cellcolor{Gray} \textbf{17.9$\pm$1.6 } \textbf{\scriptsize{\color{red}(+8.3)}} \\ \cmidrule{2-8}
			& \multirow{4}{*}{2} & FSRW~\cite{kang2019few} & 28.7$\pm$0.4&51.8$\pm$0.7&28.1$\pm$0.5&34.5$\pm$0.4&11.3$\pm$0.7 \\
			& & FRCN+ft \cite{yan2019meta} & 31.3$\pm$0.5 & 50.2$\pm$0.9 & 33.5$\pm$0.6 & 39.1$\pm$0.5 & 8.0$\pm$0.8  \\
			& & TFA  \cite{wang2020frustratingly} &41.8$\pm$0.4 & 65.6$\pm$0.6 &45.3$\pm$0.4 & \textbf{50.7$\pm$0.3}& 15.1$\pm$1.3  \\
			& &\cellcolor{Gray} {DeFRCN} & \cellcolor{Gray}\textbf{44.0$\pm$0.4} \textbf{\scriptsize{\color{red}(+2.2)}} & \cellcolor{Gray} \textbf{69.5$\pm$0.7} \textbf{\scriptsize{\color{red}(+3.9)}} & \cellcolor{Gray} \textbf{47.7$\pm$0.5} \textbf{\scriptsize{\color{red}(+2.4)}} & \cellcolor{Gray} 50.2$\pm$0.2 \textbf{\scriptsize{\color{blue}(-0.5)}} & \cellcolor{Gray} \textbf{26.0$\pm$1.3 } \textbf{\scriptsize{\color{red}(+10.9)}} \\ \cmidrule{2-8}
			& \multirow{4}{*}{3} & FSRW~\cite{kang2019few} & 
			29.2$\pm$0.4&52.7$\pm$0.6&28.5$\pm$0.4&34.2$\pm$0.3&14.2$\pm$0.7 \\
			& & FRCN+ft \cite{yan2019meta} & 32.1$\pm$0.5 & 51.3$\pm$0.8 & 34.3$\pm$0.6 & 39.1$\pm$0.5 & 11.1$\pm$0.9 \\
			& &TFA  \cite{wang2020frustratingly} & 43.1$\pm$0.4 & 67.5$\pm$0.5 & 46.7$\pm$0.5 &\textbf{51.1$\pm$0.3 }& 18.9$\pm$1.1  \\
			& &\cellcolor{Gray} {DeFRCN} & \cellcolor{Gray}\textbf{45.1$\pm$0.3} \textbf{\scriptsize{\color{red}(+2.0)}} & \cellcolor{Gray} \textbf{70.9$\pm$0.5} \textbf{\scriptsize{\color{red}(+3.4)}} & \cellcolor{Gray} \textbf{48.8$\pm$0.4} \textbf{\scriptsize{\color{red}(+2.1)}} & \cellcolor{Gray} 50.5$\pm$0.2 \textbf{\scriptsize{\color{blue}(-0.6)}} & \cellcolor{Gray} \textbf{29.2$\pm$1.0} \textbf{\scriptsize{\color{red}(+10.3)}} \\ \cmidrule{2-8}
			& \multirow{4}{*}{5} & FSRW~\cite{kang2019few} & 
			30.1$\pm$0.3&53.8$\pm$0.5&29.3$\pm$0.4&34.1$\pm$0.3&18.0$\pm$0.7 \\
			& & FRCN+ft \cite{yan2019meta} & 32.4$\pm$0.5 & 51.7$\pm$0.8 & 34.4$\pm$0.6 & 38.5$\pm$0.5 & 14.0$\pm$0.9 \\
			& & TFA  \cite{wang2020frustratingly} & 44.1$\pm$0.3 & 69.1$\pm$0.4 & 47.8$\pm$0.4 &\textbf{51.3$\pm$0.2} & 22.8$\pm$0.9  \\
			& &\cellcolor{Gray} {DeFRCN} & \cellcolor{Gray}\textbf{46.2$\pm$0.3} \textbf{\scriptsize{\color{red}(+2.1)}} & \cellcolor{Gray} \textbf{72.4$\pm$0.4} \textbf{\scriptsize{\color{red}(+3.3)}} & \cellcolor{Gray} \textbf{50.0$\pm$0.5} \textbf{\scriptsize{\color{red}(+2.2)}} & \cellcolor{Gray} 51.0$\pm$0.2 \textbf{\scriptsize{\color{blue}(-0.3)}} & \cellcolor{Gray} \textbf{32.3$\pm$0.9 } \textbf{\scriptsize{\color{red}(+9.5)}} \\ \cmidrule{2-8}
			& \multirow{3}{*}{10} & FRCN+ft \cite{yan2019meta} & 33.1$\pm$0.5 & 53.1$\pm$0.7 & 35.2$\pm$0.5 & 38.0$\pm$0.5 & 18.4$\pm$0.8  \\
			& & TFA  \cite{wang2020frustratingly} & 45.0$\pm$0.3 & 70.3$\pm$0.4 & 48.9$\pm$0.4 & \textbf{51.6$\pm$0.2} & 25.4$\pm$0.7  \\			
			& &\cellcolor{Gray} {DeFRCN} & \cellcolor{Gray}\textbf{47.0$\pm$0.3} \textbf{\scriptsize{\color{red}(+2.0)}} & \cellcolor{Gray}\textbf{ 73.3$\pm$0.3} \textbf{\scriptsize{\color{red}(+3.0)}} & \cellcolor{Gray} \textbf{51.0$\pm$0.4} \textbf{\scriptsize{\color{red}(+2.1)}} & \cellcolor{Gray} 51.3$\pm$0.2 \textbf{\scriptsize{\color{blue}(-0.3)}} & \cellcolor{Gray} \textbf{34.7$\pm$0.7} \textbf{\scriptsize{\color{red}(+9.3)}} \\
			\bottomrule
	\end{tabular}}
	\vspace{-0.18cm}
	\caption{Generalized few-shot object detection (\textit{G-FSOD}) performance on PASCAL VOC dataset. For each metric, we report the average and 95\% confidence interval computed over 30 random samples. All comparison results refer from \cite{wang2020frustratingly}.}
	\vspace{-0.0cm}
	\label{tab:gfsod-voc}
\end{table*}

\minisection{COCO.}
The Table \ref{tab:gfsod-coco} shows the \textit{G-FSOD} results on COCO dataset over $K = 1, 2, 3, 5, 10, 30$ shots. Although COCO is much more complicated than VOC, similar observations can be drawn about accuracy on both base classes and novel classes.  Concretely, the performance on base classes is comparable to TFA, but we are far superior to TFA in terms of both novel and overall results. In addition, we further notice that as the number of support shots increases, our approach can bring more performance improvements.

%--------------------------------------------------------------------------
% COCO G-FSOD Result
\begin{table*}[htb] \centering
	\setlength{\tabcolsep}{2mm}
	\scalebox{0.9}{
		\begin{tabular}{c|c|ccc|c|c}
			\toprule			
			\multirow{2}{*}{\# shots} &\multirow{2}{*}{Method} &  \multicolumn{3}{c|}{Overall \textit{\#80}}  &\multicolumn{1}{c|}{Base \textit{\#60}} & \multicolumn{1}{c}{Novel \textit{\#20}} \\ \cmidrule{3-7} 
			&  & AP & AP50 & AP75 & AP & AP \\ \midrule
			\multirow{3}{*}{1} & FRCN+ft \cite{yan2019meta} & 16.2$\pm$0.9 & 25.8$\pm$1.2 & 17.6$\pm$1.0 & 21.0$\pm$1.2 & 1.7$\pm$0.2  \\
			& TFA \cite{wang2020frustratingly} &  \textbf{24.4$\pm$0.6} & \textbf{39.8$\pm$0.8} & 26.1$\pm$0.8 &\textbf{31.9$\pm$0.7} & 1.9$\pm$0.4  \\
			&\cellcolor{Gray} {DeFRCN (Ours)} & \cellcolor{Gray} 24.0$\pm$0.4 \textbf{\scriptsize{\color{blue}(-0.4)}} & \cellcolor{Gray} 36.9$\pm$0.6  \textbf{\scriptsize{\color{blue}(-2.9)}} & \cellcolor{Gray} \textbf{26.2$\pm$0.4}  \textbf{\scriptsize{\color{red}(+0.1)}} & \cellcolor{Gray} 30.4$\pm$0.4 \textbf{\scriptsize{\color{blue}(-1.5)}} & \cellcolor{Gray} \textbf{4.8$\pm$0.6}   \textbf{\scriptsize{\color{red}(+2.9)}} \\ \midrule
			\multirow{3}{*}{2} & FRCN+ft \cite{yan2019meta} & 15.8$\pm$0.7 & 25.0$\pm$1.1 & 17.3$\pm$0.7 & 20.0$\pm$0.9 & 3.1$\pm$0.3  \\
			& TFA \cite{wang2020frustratingly} & 24.9$\pm$0.6 &\textbf{40.1$\pm$0.9} & 27.0$\pm$0.7 & \textbf{31.9$\pm$0.7}  &3.9$\pm$0.4  \\
			&\cellcolor{Gray} {DeFRCN (Ours)} & \cellcolor{Gray}\textbf{25.7$\pm$0.5} \textbf{\scriptsize{\color{red}(+0.8)}} & \cellcolor{Gray} 39.6$\pm$0.8  \textbf{\scriptsize{\color{blue}(-0.5)}} & \cellcolor{Gray} \textbf{28.0$\pm$0.5} \textbf{\scriptsize{\color{red}(+1.0)}} & \cellcolor{Gray} 31.4$\pm$0.4 \textbf{\scriptsize{\color{blue}(-0.5)}} & \cellcolor{Gray} \textbf{8.5$\pm$0.8} \textbf{\scriptsize{\color{red}(+4.6)}}  \\ \midrule
			\multirow{3}{*}{3} & FRCN+ft \cite{yan2019meta} & 15.0$\pm$0.7 & 23.9$\pm$1.2 & 16.4$\pm$0.7 & 18.8$\pm$0.9 & 3.7$\pm$0.4  \\
			& TFA \cite{wang2020frustratingly} &25.3$\pm$0.6 & 40.4$\pm$1.0 & 27.6$\pm$0.7 & 32.0$\pm$0.7 & 5.1$\pm$0.6  \\
			&\cellcolor{Gray} {DeFRCN (Ours)} & \cellcolor{Gray}\textbf{26.6$\pm$0.4} \textbf{\scriptsize{\color{red}(+1.3)}} & \cellcolor{Gray} \textbf{41.1$\pm$0.7} \textbf{\scriptsize{\color{red}(+0.7)}} & \cellcolor{Gray} \textbf{28.9$\pm$0.4} \textbf{\scriptsize{\color{red}(+1.3)}}& \cellcolor{Gray} \textbf{32.1$\pm$0.3}  \textbf{\scriptsize{\color{red}(+0.1)}} & \cellcolor{Gray} \textbf{10.7$\pm$0.8} \textbf{\scriptsize{\color{red}(+5.6)}}\\ \midrule
			\multirow{3}{*}{5} & FRCN+ft  \cite{yan2019meta}& 14.4$\pm$0.8 & 23.0$\pm$1.3 & 15.6$\pm$0.8 & 17.6$\pm$0.9 & 4.6$\pm$0.5  \\
			& TFA \cite{wang2020frustratingly} & 25.9$\pm$0.6 & 41.2$\pm$0.9 &28.4$\pm$0.6 & 32.3$\pm$0.6 & 7.0$\pm$0.7 \\
			&\cellcolor{Gray} {DeFRCN (Ours)} & \cellcolor{Gray}\textbf{27.8$\pm$0.3} \textbf{\scriptsize{\color{red}(+1.9)}}& \cellcolor{Gray} \textbf{43.0$\pm$0.6} \textbf{\scriptsize{\color{red}(+1.8)}}& \cellcolor{Gray} \textbf{30.2$\pm$0.3} \textbf{\scriptsize{\color{red}(+1.8)}}& \cellcolor{Gray} \textbf{32.6$\pm$0.3} \textbf{\scriptsize{\color{red}(+0.3)}} & \cellcolor{Gray} \textbf{13.6$\pm$0.7} \textbf{\scriptsize{\color{red}(+6.6)}} \\ \midrule
			\multirow{3}{*}{10} & FRCN+ft \cite{yan2019meta} & 13.4$\pm$1.0 & 21.8$\pm$1.7 & 14.5$\pm$0.9 & 16.1$\pm$1.0 & 5.5$\pm$0.9 \\
			& TFA \cite{wang2020frustratingly} & 26.6$\pm$0.5 &42.2$\pm$0.8 & 29.0$\pm$0.6 & 32.4$\pm$0.6 & 9.1$\pm$0.5  \\			
			&\cellcolor{Gray} {DeFRCN (Ours)} & \cellcolor{Gray}\textbf{29.7$\pm$0.2} \textbf{\scriptsize{\color{red}(+3.1)}} & \cellcolor{Gray} \textbf{46.0$\pm$0.5} \textbf{\scriptsize{\color{red}(+3.8)}} & \cellcolor{Gray} \textbf{32.1$\pm$0.2} \textbf{\scriptsize{\color{red}(+3.1)}} & \cellcolor{Gray} \textbf{34.0$\pm$0.2} \textbf{\scriptsize{\color{red}(+1.6)}} & \cellcolor{Gray} \textbf{16.8$\pm$0.6} \textbf{\scriptsize{\color{red}(+7.7)}}  \\ \midrule
			\multirow{3}{*}{30} & FRCN+ft \cite{yan2019meta} & 13.5$\pm$1.0 & 21.8$\pm$1.9 & 14.5$\pm$1.0 & 15.6$\pm$1.0 & 7.4$\pm$1.1  \\
			& TFA \cite{wang2020frustratingly} & 28.7$\pm$0.4 &44.7$\pm$0.7 & 31.5$\pm$0.4 & 34.2$\pm$0.4 &12.1$\pm$0.4  \\
			&\cellcolor{Gray} {DeFRCN (Ours)} & \cellcolor{Gray}\textbf{31.4$\pm$0.1} \textbf{\scriptsize{\color{red}(+2.7)}} & \cellcolor{Gray} \textbf{48.8$\pm$0.2} \textbf{\scriptsize{\color{red}(+4.1)}} & \cellcolor{Gray} \textbf{33.9$\pm$0.1} \textbf{\scriptsize{\color{red}(+2.4)}} & \cellcolor{Gray} \textbf{34.8$\pm$0.1 } \textbf{\scriptsize{\color{red}(+0.6)}} & \cellcolor{Gray} \textbf{21.2$\pm$0.4 } \textbf{\scriptsize{\color{red}(+9.1)}} \\
			\bottomrule
	\end{tabular}}
	\vspace{-0.18cm}
	\caption{Generalized few-shot object detection (\textit{G-FSOD}) performance on COCO dataset. For each metric, we report the average and 95\% confidence interval computed over 10 random samples. All comparison results refer from \cite{wang2020frustratingly}. }
	\vspace{-0.0cm}
	\label{tab:gfsod-coco}
\end{table*}
%--------------------------------------------------------------------------

\section{Additional Analysis on PCB}
\subsection{Boost Other Approaches with PCB}

As a plug-and-play module, the proposed PCB is easily equipped to any other architectures to build stronger few-shot detectors. Here, we verify this argument with introducing PCB into other previous approaches, including FRCN-ft \cite{yan2019meta}, TFA \cite{wang2020frustratingly}, MPSR \cite{wu2020multi}, and all experimental results on COCO dataset are shown in the Table \ref{tab:pcb-plug-and-play}. Regardless of methods or the number of shots, we observe that using PCB  can consistently achieve much higher performance (+1.0$\%$ $\sim$ +3.0$\%$ points) on novel classes, which demonstrates the effectiveness and flexibility of our PCB module.

%--------------------------------------------------------------------------
\begin{table*}[htb] \centering
	\setlength{\tabcolsep}{2.5mm}
	\scalebox{1.0}{
		\begin{tabular}{c|l|l|l|c|cccccc}
			%\hline
			\toprule[1.1pt]
			\multicolumn{4}{c|}{}           & \multicolumn{1}{c|}{}               & \multicolumn{6}{c}{\# shots }                         \\
			\multicolumn{4}{c|}{\multirow{-2}{*}{Method}}  & \multicolumn{1}{c|}{\multirow{-2}{*}{w / PCB}} & 1        & 2              & 3             & 5             & 10    &30   \\ \midrule[0.9pt]
			\multicolumn{4}{c|}{}   &\xmark & 1.0    & 1.8     & 2.8          & 4.0          & 6.9          &11.0    \\
			\multicolumn{4}{c|}{\multirow{-2}{*}{FRCN-ft \cite{yan2019meta}}}   &\cmark & 2.4 \textbf{\scriptsize{\color{red}(+1.4)}}     & 4.1 \textbf{\scriptsize{\color{red}(+2.3)}}    & 5.2 \textbf{\scriptsize{\color{red}(+2.4)}}         & 6.6  \textbf{\scriptsize{\color{red}(+2.6)}}        & 9.9  \textbf{\scriptsize{\color{red}(+3.0)}}        & 14.0 \textbf{\scriptsize{\color{red}(+3.0)}}        \\
			\midrule[0.9pt]
			\multicolumn{4}{c|}{}   &\xmark     & 4.4  &5.4       &6.0          & 7.7          &9.0         & 13.4     \\
			\multicolumn{4}{c|}{\multirow{-2}{*}{TFA  \cite{wang2020frustratingly}}}   &\cmark     & 6.7  \textbf{\scriptsize{\color{red}(+2.3)}}   & 7.6 \textbf{\scriptsize{\color{red}(+2.2)}}    & 9.0 \textbf{\scriptsize{\color{red}(+3.0)}}         & 10.4 \textbf{\scriptsize{\color{red}(+2.7)}}         & 11.8 \textbf{\scriptsize{\color{red}(+2.8)}}         & 15.5 \textbf{\scriptsize{\color{red}(+2.1)}}             \\
			\midrule[0.9pt]
			\multicolumn{4}{c|}{}  &\xmark        &5.1  &6.7       & 7.4         & 8.7       &9.8         & 14.5        \\
			\multicolumn{4}{c|}{\multirow{-2}{*}{MPSR  \cite{wu2020multi}}}   &\cmark      & 6.7  \textbf{\scriptsize{\color{red}(+1.6)}}   & 8.9 \textbf{\scriptsize{\color{red}(+2.2)}}    & 9.7 \textbf{\scriptsize{\color{red}(+2.3)}}         & 10.9 \textbf{\scriptsize{\color{red}(+2.2)}}         & 11.9 \textbf{\scriptsize{\color{red}(+2.1)}}         & 15.5 \textbf{\scriptsize{\color{red}(+1.0)}}       \\
			\midrule[0.9pt]
			\multicolumn{4}{c|}{}    &\xmark      & 7.9 & 10.9 & 13.4 & 14.6 & 16.9 & 21.0  \\ 
			\multicolumn{4}{c|}{\multirow{-2}{*}{DeFRCN (Ours)}}    &\cmark      & 9.3 \textbf{\scriptsize{\color{red}(+1.4)}} &12.9 \textbf{\scriptsize{\color{red}(+2.0)}} & 14.8 \textbf{\scriptsize{\color{red}(+1.4)}} & 16.1 \textbf{\scriptsize{\color{red}(+1.5)}} &18.5 \textbf{\scriptsize{\color{red}(+1.6)}} & 22.6 \textbf{\scriptsize{\color{red}(+1.6)}}  \\ 
			\bottomrule[1.1pt]
	\end{tabular}}
	\vspace{-0.18cm}
	\caption{Effectiveness of Prototypical Calibration Block with different approaches. We evaluate \textit{FSOD} performance ($mAP$) on COCO dataset with $K=1,2,3,5,10,30$ shots over multiple runs. All experimental results are reproduced by us. The term w/PCB indicates whether the method uses the PCB module. Note that the $\alpha$ in PCB is set to 0.5 in all experiments.}
	\vspace{-0.0cm}
	\label{tab:pcb-plug-and-play}
\end{table*}
%--------------------------------------------------------------------------

\subsection{Employ Other Pre-trained Models}
In the main paper, we utilize the standard ImageNet pre-trained model (IN-1K), which is widely adopted in most of few-shot object detection frameworks, to initialize both Faster-RCNN and PCB. Since the core module of PCB is the generalizable feature extractor, which determines the final performance of the score calibration, we further explore other pre-trained models (see Table \ref{tab:pretrain}) in this section. SwAV \cite{caron2020unsupervised} is an efficient method for pre-training without using annotations, $\ie$, self-supervised learning. IN-SwAV indicates that the model is pre-trained by SwAV on ImageNet. IG-WSL \cite{wslimageseccv2018} employs the ResNeXt \cite{xie2017aggregated} architecture and pre-trains on a much larger social media image dataset (Instagram) with weakly-supervised learning paradigm.  Table \ref{tab:pcb-pretrain-voc} shows the performance on VOC with utilizing the above three pre-trained models. No matter which one is exploited, the final performance is better with PCB. Moreover, we further notice that using a stronger pre-trained model, the performance of FSOD can be improved more.

%--------------------------------------------------------------------------
% PCB + Other approach
\begin{table}[H] \centering
	\setlength{\tabcolsep}{1.1mm}
	\vspace{-0.05cm}
	\scalebox{0.85}{
		\begin{tabular}{c|c|c|c|c}
			\toprule[1.1pt]
			Method & Backbone & Paradigm & \# Images & \# Classes \\
			\midrule[0.9pt]
			IN-SwAV \cite{caron2020unsupervised} & ResNet-50 & S-S-L & 1.28M & 0 \\
			IN-1K \cite{he2016deep} & ResNet-101 & S-L & 1.28M & 1000 \\
			IG-WSL \cite{wslimageseccv2018} & ResNeXt-101 & W-S-L& 940M & 1000 \\
			\bottomrule[1.1pt]
	\end{tabular}}
	\vspace{-0.18cm}
	\caption{
		The comparison between different pre-trained classification models. S-S-L, S-L and W-S-L stand for self-supervised learning, supervised learning and weakly-supervised learning respectively.}
	\vspace{-0.0cm}
	\label{tab:pretrain}
\end{table}
%--------------------------------------------------------------------------

%--------------------------------------------------------------------------
% PCB + Other approach
\begin{table*}[htb] \centering
	\setlength{\tabcolsep}{1.4mm}
	\scalebox{0.94}{
		\begin{tabular}{c|l|l|l|c|ccccc|ccccc|ccccc}
			%\hline
			\toprule[1.1pt]
			\multicolumn{4}{c|}{}    & \multicolumn{1}{c|}{}        & \multicolumn{5}{c|}{Novel Set 1}                                              & \multicolumn{5}{c|}{Novel Set 2}                                              & \multicolumn{5}{c}{Novel Set 3}                                              \\
			\multicolumn{4}{c|}{\multirow{-2}{*}{Method}} & \multicolumn{1}{c|}{\multirow{-2}{*}{Model}} & 1             & 2             & 3             & 5             & 10            & 1             & 2             & 3             & 5             & 10            & 1             & 2             & 3             & 5             & 10            \\ \midrule[0.9pt]
			\multicolumn{4}{c|}{} &\xmark & 47.0 & 48.8 & 52.3 & 57.1 & 55.6 & 22.5& 31.9& 42.1& 45.6 & 42.3& 42.5&48.7& 48.9  &  51.1& 52.2 \\ 
			\cmidrule[0.9pt]{5-20}
			\multicolumn{4}{c|}{} & IN-SwAV \cite{caron2020unsupervised} & 48.7 & 52.4 &54.5 & 60.2 & 56.3 & 26.9& 34.6& 44.6& 48.1 & 44.7& 41.8&50.1&50.5&  53.4& 55.1 \\ 
			\multicolumn{4}{c|}{} & IN-1K \cite{he2016deep} & 53.6 & 57.5 & 61.5 & 64.1 & 60.8 & 30.1& 38.1& 47.0& 53.3 & 47.9& 48.4&50.9& 52.3 &  54.9& 57.4 \\ 
			\multicolumn{4}{c|}{\multirow{-4}{*}{DeFRCN}} & IG-WSL \cite{wslimageseccv2018} & 62.3 & 64.5 & 66.6 & 69.3 & 68.2 & 37.5& 44.7& 53.5& 57.6 & 54.7& 54.7&57.2& 59.0 &  60.9& 62.0 \\ 
			\bottomrule[1.1pt]
	\end{tabular}}
	\vspace{-0.18cm}
	\caption{
		Experimental results of employing different pre-trained model in PCB on PASCAL VOC dataset. All reported results are averaged over 30 random samples. IN-SwAV, IN-1K and INS-WSL denote the different pre-trained models from ImageNet self-supervised learning, conventional  supervised learning and weakly-supervised learning separately.}
	\vspace{-0.0cm}
	\label{tab:pcb-pretrain-voc}
\end{table*}
%--------------------------------------------------------------------------

\subsection{Why PCB Works ? }
\begin{figure*}[htb]
	\vspace{-0.cm}
	\begin{center}
		\includegraphics[width=1.0\linewidth]{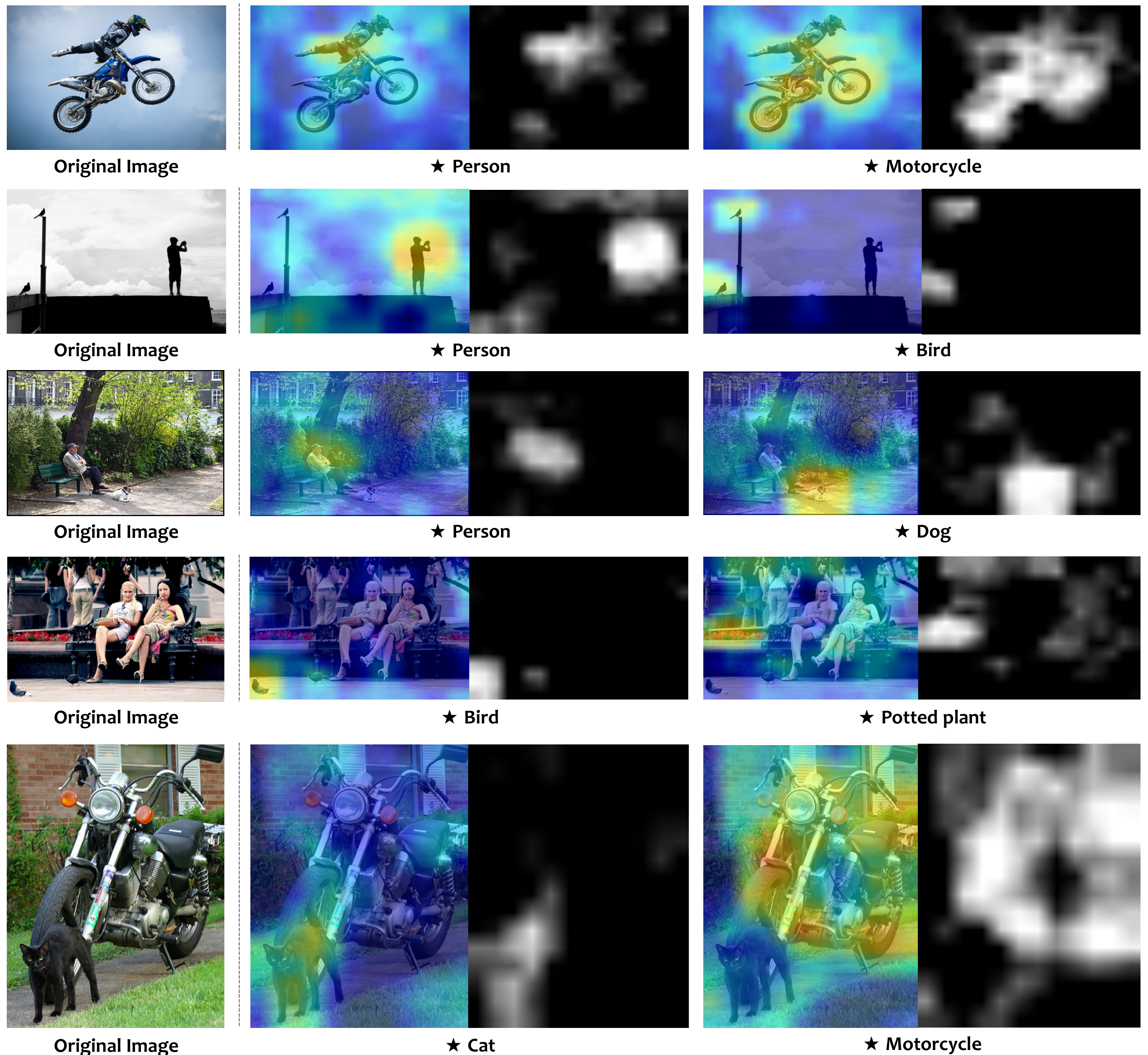}
	\end{center}
	\vspace{-0.2cm}
	\caption{The visualization of PCB on COCO val set. Through different kinds of prototypes, which are calculated by $K$-shot samples ($K=10$), distinct areas of the same picture are activated. The symbol $\bigstar$ indicates that it is some kind of prototypes.}
	\vspace{-0.4cm}
	\label{fig:heatmap}
\end{figure*}

The PCB can be reinterpreted as a non-parameter few-shot classification model, which draws on the idea of Prototypical Network \cite{Snell2017}. Based on the COCO 10-shot task, we calculate the channel-wise cosine similarity between different few-shot RoI prototypes ($C \times 1 \times 1$) and the feature map ($C \times H \times W$) of the test image, and then visualize the similarity map in Fig.\ref{fig:heatmap}. We find that the prototypes from different categories can indeed activate distinct areas of the feature map, which indicates that the metric-based pairwise score in data-scarce scenario is very effective. In addition, we notice that even if the category label of novel prototype is not seen before by the pre-trained classification model,  an ideal similarity map can still be obtained, $\eg$, the novel label 'Person' does not exist in ImageNet 1K sysnets, see the first three lines in Fig.\ref{fig:heatmap}.  Moreover, the results of IN-SwAV ($\ie$ self-supervised paradigm) in Table \ref{tab:pcb-pretrain-voc} further prove this argument. According to the visualization and above analysis, we believe that it is reasonable for PCB to utilize the pairwise score based on classification  model to calibrate the softmax score from the original classification branch of few-shot detector.
%--------------------------------------------------------------------------
%--------------------------------------------------------------------------

\section{Related extensions of GDL}

\subsection{Conventional Cross-Domain Object Detection}
In the experimental section of the main paper, we have verified that the proposed GDL is not only remarkably effective for few-shot object detection ( $\ie$, \textit{FSOD}, \textit{G-FSOD} and cross-domain \textit{FSOD}), but also plays a positive role in conventional object detection. In this section, we further explore the conventional cross-domain object detection and all experimental results are shown in Table \ref{tab:cross-domain-conv}. We use the Cityscapes \cite{cordts2016cityscapes} and FoggyCityscapes \cite{sakaridis2018semantic} (Normal-to-Foggy) as our benchmarks and follow the same evaluation protocol in \cite{zheng2020cross}. By comparing the experimental results of the second row and the third row in Table \ref{tab:cross-domain-conv}, we find that adding GDL achieves 32.8\% mAP on the weather transfer task, which is +2.8\% higher than the plain Faster-RCNN.

%--------------------------------------------------------------------------
\begin{table*}[htb] \centering
	\setlength{\tabcolsep}{2mm}
	\scalebox{0.8}
	{\begin{tabular}{c|l|l|l|c|cccccccc|c}
			%\hline
			\toprule[1.1pt]
			\multicolumn{14}{c}{Cityscapes $\to$ FoggyCityscapes} \\
			\midrule[0.9pt]
			\multicolumn{4}{c|}{Method} & Backbone  &  Bus    & Bicycle & Car &Motor & Person& Rider&Train&Truck &$mAP$ \\ 
			\midrule[0.9pt]
			\multicolumn{4}{c|}{Faster-RCNN \cite{zheng2020cross}} & VGG16  &  25.0    & 26.8 & 30.6 &15.5 & 24.1& 29.4&4.6&10.6 & 20.8 \\
			\midrule[0.9pt]
			\multicolumn{4}{c|}{Faster-RCNN $^\ast$} & ResNet-101  & 31.5    &\textbf{ 39.3} & 45.2 &24.7 & \textbf{35.3}& 41.2&8.8&18.7& 30.0 \\
			\midrule[0.9pt]
			\multicolumn{4}{c|}{+ GDL} & ResNet-101 &  \textbf{32.9}  \textbf{\scriptsize{\color{red}(+1.4)}}  & 38.4 \textbf{\scriptsize{\color{blue}(-0.9)}} & \textbf{47.3} \textbf{\scriptsize{\color{red}(+2.1)}}  &\textbf{26.6} \textbf{\scriptsize{\color{red}(+1.9)}} & 34.3 \textbf{\scriptsize{\color{blue}(-1.0)}} & \textbf{41.4} \textbf{\scriptsize{\color{red}(+0.2)}} &\textbf{17.3} \textbf{\scriptsize{\color{red}(+8.5)}} &\textbf{24.3} \textbf{\scriptsize{\color{red}(+7.6)}}& \textbf{32.8} \textbf{\scriptsize{\color{red}(+2.8)}} \\
			\bottomrule[1.1pt]
	\end{tabular}}
	\vspace{-0.18cm}
	\caption{The performance of conventional cross-domain object detection. All results in the first line refer from \cite{zheng2020cross} for brief comparison. Note that the Faster R-CNN model trained on the source domain only without any other information (denoted as “Source Only” in other papers). The symbol $^\ast$ indicates the model is re-implemented by us.}
	\vspace{-0.0cm}
	\label{tab:cross-domain-conv}
\end{table*}
%--------------------------------------------------------------------------

\subsection{The value range of $\lambda$}
We discuss the value range of $\lambda$ into three situations.
\begin{itemize}[leftmargin=9pt,topsep=1pt, parsep=-2pt, partopsep=-2pt]
	
	\item \textit{$\lambda_{rpn} \in [0, 1]$ and $\lambda_{rcnn} \in [0, 1]$}. 
	This setting has been explored in our paper and achieved the best results. 
	
	\item \textit{$\lambda_{rpn} \in (-\infty, 0)$ or $\lambda_{rcnn} \in (-\infty, 0)$}.  
	$\lambda < 0$ means that the downstream module has a negative effect on the optimization direction of backbone. Without any adversarial strategy, this setup is meaningless for object detection.
	
	\item \textit{$\lambda_{rpn} \in (1, +\infty)$ or $\lambda_{rcnn} \in (1, +\infty)$}.  
	$\lambda > 1$  means that the gradient from the downstream module magnifies its effect on the backbone. We notice that slightly increasing $\lambda$ (\eg $1 \sim 5$) will not affect the stability of detector but incite performance degradation, which is caused by the backbone's update speed faster than before and over-fitting. When $\lambda$ is relatively large (\eg $ > 5$), due to over-emphasizing the degree of coupling between the module and the backbone, the model will usually converge to an unreasonable saddle point and cause a collapse solution. The value of 5 is obtained by experiments approximately.
	
\end{itemize}

\section{More Visualization of Our Approach}
\begin{figure*}[htb]
	\vspace{-0.cm}
	\begin{center}
		\includegraphics[width=1.0\linewidth]{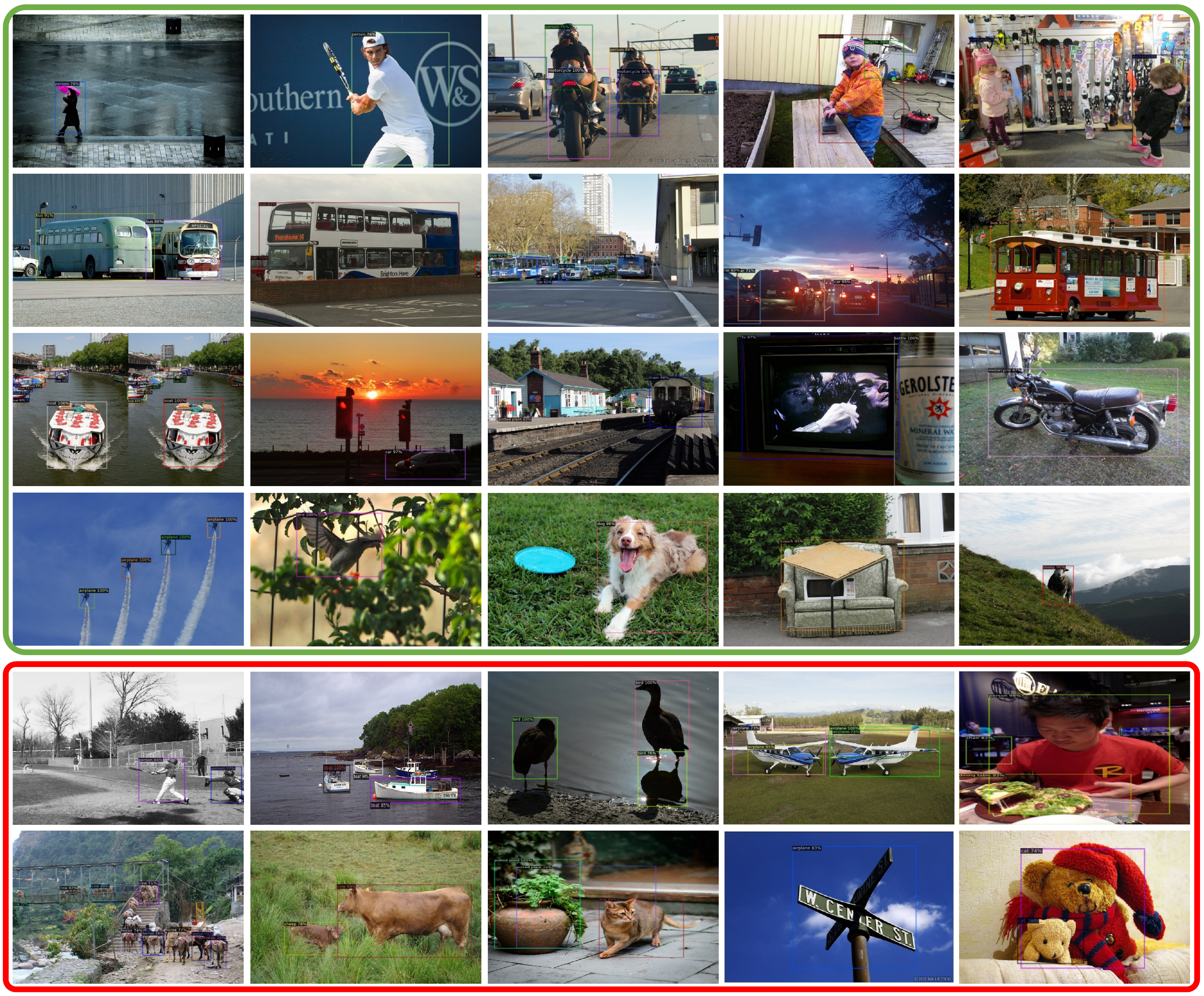}
	\end{center}
	\vspace{-0.2cm}
	\caption{The visualization results of our 10-shot object detection on COCO dataset. We visualize the bounding boxes with score larger than 0.7. The green and red box shows the success and failure cases of our DeFRCN respectively.}
	\vspace{-0.4cm}
	\label{fig:visualization}
\end{figure*}

We provide qualitative visualizations of the detected novel objects on COCO dataset in Fig.\ref{fig:visualization}. We show both success (green box) and failure cases (red box) when detecting novel objects for each image to help analyze the possible error types, including misclassifying novel objects, mislocalizing objects and missing detections.

%--------------------------------------------------------------------------

\clearpage
\clearpage
{\small
	\bibliographystyle{ieee_fullname}
	\bibliography{egbib}
}

\end{document}